\PassOptionsToPackage{table,xcdraw,prologue}{xcolor} 

\documentclass[sigconf]{acmart}

\usepackage{xcolor} 

\definecolor{myblue}{RGB}{221,242,255}
\definecolor{mygreen}{RGB}{237,247,235}
\definecolor{myye}{RGB}{255,249,230}
\definecolor{mygray}{RGB}{248,248,249}

\usepackage{pifont}
\usepackage{mathtools}
\usepackage{soul}
\usepackage{booktabs}
\usepackage{multirow}
\usepackage{makecell}
\usepackage{bm}
\usepackage{graphicx}
\usepackage{subfigure}
\usepackage{algorithm} 
\usepackage{algpseudocode}
\usepackage{amsthm}
\usepackage{multicol}
\usepackage{balance}

\newtheorem{theorem}{Theorem}
\newtheorem{assumption}{Assumption}
\newcommand{\mathcolorbox}[2]{\colorbox{#1}{$\displaystyle #2$}}

\copyrightyear{2025}
\acmYear{2025}
\setcopyright{acmlicensed}
\acmConference[KDD '25] {Proceedings of the 31st ACM SIGKDD Conference on Knowledge Discovery and Data Mining V.2}{August 3--7, 2025}{Toronto, ON, Canada.}
\acmBooktitle{Proceedings of the 31st ACM SIGKDD Conference on Knowledge Discovery and Data Mining V.2 (KDD '25), August 3--7, 2025, Toronto, ON, Canada}
\acmISBN{979-8-4007-1454-2/25/08}
\acmDOI{10.1145/3711896.3736957}
\copyrightyear{2025}
\acmYear{2025}
\setcopyright{rightsretained}
\acmConference[KDD '25] {Proceedings of the 31st ACM SIGKDD Conference on Knowledge Discovery and Data Mining V.2}{August 3--7, 2025}{Toronto, ON, Canada.}
\acmBooktitle{Proceedings of the 31st ACM SIGKDD Conference on Knowledge Discovery and Data Mining V.2 (KDD '25), August 3--7, 2025, Toronto, ON, Canada}
\acmISBN{979-8-4007-1454-2/25/08}
\acmDOI{10.1145/3711896.3736957}
\copyrightyear{2025}
\acmYear{2025}
\setcopyright{cc}
\setcctype{by}
\acmConference[KDD '25]{Proceedings of the 31st ACM SIGKDD Conference on Knowledge Discovery and Data Mining V.2}{August 3--7, 2025}{Toronto, ON, Canada}
\acmBooktitle{Proceedings of the 31st ACM SIGKDD Conference on Knowledge Discovery and Data Mining V.2 (KDD '25), August 3--7, 2025, Toronto, ON, Canada}
\acmDOI{10.1145/3711896.3736957}
\acmISBN{979-8-4007-1454-2/2025/08}

\begin{document}

\title{FedSC: Federated Learning with Semantic-Aware Collaboration}

\author{Huan Wang}
\authornote{These authors contributed equally to this research.}
\affiliation{
  \department{School of Computing and Information Technology}
  \institution{University of Wollongong}
  \city{Wollongong}
  \state{NSW}
  \country{Australia}
}
\email{hw226@uowmail.edu.au}

\author{Haoran Li}
\authornotemark[1]
\affiliation{
  \department{School of Computing and Information Technology}
  \institution{University of Wollongong}
  \city{Wollongong}
  \state{NSW}
  \country{Australia}
}
\email{hl644@uowmail.edu.au}

\author{Huaming Chen}
\affiliation{
    \department{School of Electrical and Computer Engineering}
    \institution{The University of Sydney}
    \city{Sydney}
    \state{NSW}
    \country{Australia}
}
\email{huaming.chen@sydney.edu.au}

\author{Jun Yan}
\affiliation{
  \department{School of Computing and Information Technology}
  \institution{University of Wollongong}
  \city{Wollongong}
  \state{NSW}
  \country{Australia}
}
\email{jyan@uow.edu.au}

\author{Jiahua Shi}
\affiliation{
  \department{QLD Alliance for Agriculture and Food Innovation}
  \institution{The University of Queensland}
  \city{Brisbane}
  \state{QLD}
  \country{Australia}
}
\email{jiahua.shi@uq.edu.au}

\author{Jun Shen}
\authornote{Jun Shen is the corresponding author.}
\affiliation{
  \department{School of Computing and Information Technology}
  \institution{University of Wollongong}
  \city{Wollongong}
  \state{NSW}
  \country{Australia}
}
\email{jshen@uow.edu.au}

\renewcommand{\shortauthors}{Huan Wang et al.}

\begin{abstract}
Federated learning (FL) aims to train models collaboratively across clients without sharing data for privacy-preserving. However, one major challenge is the data heterogeneity issue, which refers to the biased labeling preferences at multiple clients. A number of existing FL methods attempt to tackle data heterogeneity locally (\textit{e.g.}, regularizing local models) or globally (\textit{e.g.}, fine-tuning global model), often neglecting inherent semantic information contained in each client. 
To explore the possibility of using intra-client semantically meaningful knowledge in handling data heterogeneity, in this paper, we propose \textbf{Fed}erated Learning with \textbf{S}emantic-Aware \textbf{C}ollaboration (FedSC) to capture client-specific and class-relevant knowledge across heterogeneous clients. 
The core idea of FedSC is to construct relational prototypes and consistent prototypes at semantic-level, aiming to provide fruitful class underlying knowledge and stable convergence signals in a prototype-wise collaborative way. 
On the one hand, FedSC introduces an inter-contrastive learning strategy to bring instance-level embeddings closer to relational prototypes with the same semantics and away from distinct classes. 
On the other hand, FedSC devises consistent prototypes via a discrepancy aggregation manner, as a regularization penalty to constrain the optimization region of the local model. 
Moreover, a theoretical analysis for FedSC is provided to ensure a convergence guarantee. 
Experimental results on various challenging scenarios demonstrate the effectiveness of FedSC and the efficiency of crucial components. 
Our code is at \href{https://github.com/hwang52/FedSC}{https://github.com/hwang52/FedSC}.
\end{abstract}



\maketitle

\begin{figure}[t!]
    \centering
    \includegraphics[width=0.9\linewidth]{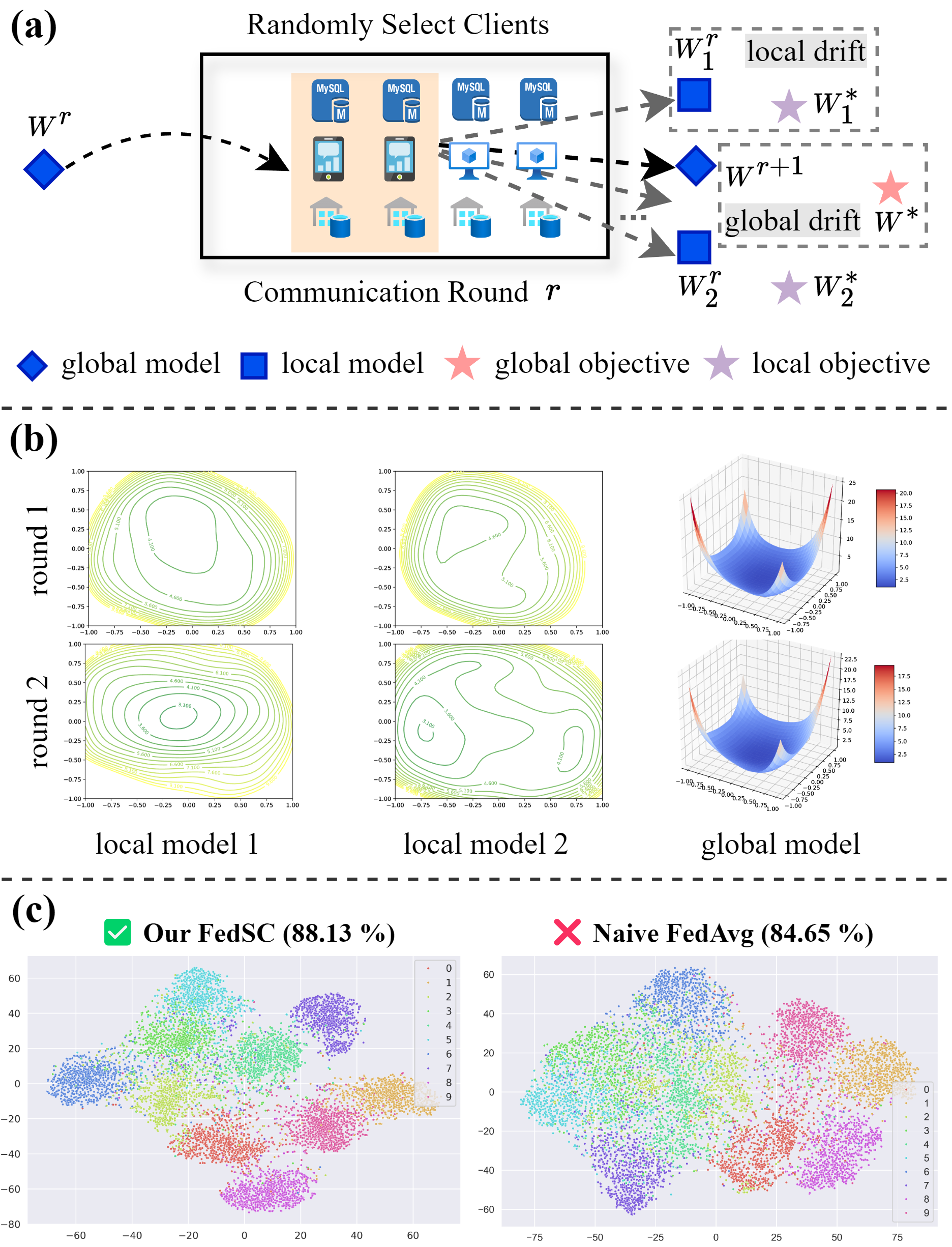}
    \caption{(a) An example of local and global drift in FL training on round $r$; (b) Loss landscape visualization of the local models and global model, the vertical axis shows the loss; (c) The performance of FedSC and FedAvg on CIFAR10 dataset with \textit{non-iid} distribution shows that our method promotes a more generalizable FL global model.}
    \label{fig:fig1}
\end{figure}

\section{Introduction}
Federated Learning (FL) paradigm involves the collaborative training of a global model by periodically aggregating model parameters from multiple privacy-preserving decentralized clients~\cite{McMahan17,li2020fedprox}. It enables clients to collectively train a global model without sharing private data, facilitating different real-world applications in multiple branches (\textit{e.g.}, autonomous driving~\cite{9827020} and recommendation system~\cite{yu2023untargeted}). 
In the standard FL setting, each client trains a local model using their private data, sends the model updates to a central server for weight aggregation, and subsequently updates their local model with the new global model parameters received from the server. 
Despite the promising advancements made by FL in private data modeling, non-independent and identically distributed (\textit{non-iid}) data is still an outstanding challenge drawing a lot of attention~\cite{9835537,kairouz2021advances,li2020federated,tan2023federated,ZHU2021371}. 
Such \textit{non-iid} data distribution typically occurs in FL, which introduces drift in both local (client) and global (server) model optimizations, thus making the convergence slow and unstable~\cite{pmlr-v119-karimireddy20a,li2020fedprox,yuan2022convergence} during FL training across clients.

To alleviate data heterogeneity, there are mainly two research directions: 1) one direction focuses on stabilizing local training by mitigating the deviation between local models and the global model within the parameter space to counteract local drift~\cite{pmlr-v119-karimireddy20a,gao2022federated,li2020fedprox}; 2) another solution involves the server to enhance the effectiveness of global weight aggregation~\cite{Wang2020Federated,nguyen2022federated,3485730}. 
However, data heterogeneity is intrinsic to both local and global drifts, and these methods often address only one part of this issue. Therefore, we speculate that naive learning on the private data in FL brings poor generalizable ability: \textit{local private model overfits local domain data distribution}, which is magnified under the data heterogeneity and thus hinders the improvement of generalizability and discriminability. 
Fig.~\ref{fig:fig1} (a) provides an intuitive illustration of the data heterogeneity with \textit{non-iid} clients in round $r$. Specifically, each client trains locally on their private data with the initial model $w^{r}$ towards their respective local model convergence point (\textit{e.g.}, $w^{r}_{1}$ and $w^{r}_{2}$). However, these local minima may not align well with the local objective (\textit{e.g.}, $w^{*}_{1}$ and $w^{*}_{2}$), introducing drift in clients' local updates. Additionally, the global model $w^{r+1}$ learned through aggregation at the server deviates from the optimal point $w^{*}$. The straightforward reason for this deviation is that the local models overfit to their local convergence point, leading to an insufficient generalization of the global model. We also conduct a preliminary experiment based on loss landscape visualization~\cite{visualloss}. In Fig.~\ref{fig:fig1} (b), the local convergence points of different clients are varied even occurring within the same round. Since each client has its own loss landscape, the aggregated global model strays from the global objective during optimization. In fact, most of the previous FL works force the local model to be consistent with the global model. Although they demonstrate a certain effect by reducing the data heterogeneity locally or globally, the gradually enlarged parameter deviation persists.

Taking into account both the effectiveness and efficiency in FL, we revisit the efforts on the class-wise prototype~\cite{wei2023online,zhu2021prototype,yang2018robust,li2021adaptive,deng2021variational,tan2022fedproto}, which is defined as the mean value of features with identical semantics. It represents class-relevant semantic characteristics and can be an effective information carrier for communication between the clients and server. The compelling representative semantic and exchangeable proxy abilities of class-wise prototypes motivate us to rethink: \textbf{how to effectively unleash semantically meaningful knowledge via prototypes to promote FL training?}

However, there are two main challenges in solving this question: 1) the prototypes on a single client would struggle to present learned semantic representations due to the inherent heterogeneity in FL; 2) directly averaging all prototypes across clients to get global prototypes would bring the same impediment as global aggregation, as simply averaging operation weakens the domain diversity and semantic relations with different classes. Besides, in the long-tail distribution scenario with significant class imbalance, global prototypes would likely yield bias to the semantic features of head classes and ignore the tail classes, resulting in poor performance. 
Driven by these two issues, we construct relational prototypes and consistent prototypes at semantic-level: 
\textbf{1)} On the one hand, we argue that semantic knowledge is better presented by mutual relations of the prototypes across clients than a single client-specific prototype. Therefore, we combine the semantic relations of different clients' class prototypes and design \textbf{relational prototypes} by penalizing the angular differences. Since an angle is a higher-order property than a distance, we ensemble the clients' prototypes based on the angular relationships, which helps capture rich domain variance and class-relevant information. 
\textbf{2)} On the other hand, we generate \textbf{consistent prototypes} by merging the relational prototypes via a discrepancy aggregation way to construct stable global signals, avoiding over-optimization and ensuring convergence stability. 
We will analyze the superiority of relational and consistent prototypes and how to construct them in Sec.~\ref{sec:rep} and Sec.~\ref{sec:cop}.

Motivated by these insights, in this paper, we propose a novel \textbf{Fed}erated Learning with \textbf{S}emantic-Aware \textbf{C}ollaboration (FedSC) to alleviate the challenges posed by data heterogeneity in FL, which consists of two key components: \textbf{1) First}, to promote the generalizability and discriminability of local training, we introduce Relational Prototypes Contrastive Learning (RPCL), which utilizes relational prototypes to devise an inter-contrastive learning~\cite{chen2020simple,he2020momentum,zhuang2020comprehensive,le2020contrastive} strategy. By maximizing the agreement between query embedding and its relational prototype, each client effectively captures class-relevant information on their specific domain. RPCL adaptively encourages instance embedding to be more similar to relational prototypes from the same class than other relational prototypes with different semantics. Such a strategy implicitly pulls samples with the same semantics closer together and separates other classes away in the feature space, maintaining a clear decision boundary. \textbf{2) Second}, we leverage consistent prototypes to construct a flat and stable convergent target, and develop Consistent Prototypes Discrepancy Regularization (CPDR). Specifically, we combine the relational prototypes to acquire consistent prototypes via a discrepancy aggregation manner rather than simply averaging, which can determine more distinguishing weights for representative clients. The discrepancy is defined as the distance between local category distribution and hypothetical global category distribution over all clients. The local instance is required to align the representations with the corresponding consistent prototype on particular semantics. Therefore, the local model would not be biased toward local domain and exhibit stable performance. 
Benefiting from these two components, FedSC effectively unleashes the potential of semantic knowledge among clients, making the FL global model converge quickly and reach better performance, shown in Fig.~\ref{fig:fig1} (c).
\begin{itemize}
  \item We focus on excavating and transferring the semantically meaningful knowledge across clients. Inspired by the success of prototypes, we further devise relational and consistent prototypes, capturing rich class-relevant domain diversity and providing stable global signals.
  \item We propose FedSC, a concise yet effective FL method that alleviates the negative impact of data heterogeneity by two key components: RPCL enhances the local model to learn more general semantic representations; CPDR enforces the uniformity of the feature space for identical semantics.
  \item We provide a theoretical analysis of our method under non-convex objectives, ensuring convergence and a guaranteed convergence rate with detailed derivation (Sec.~\ref{sec:ana}).
  \item We conduct extensive experiments in challenging settings (\textit{e.g.}, label shift, long-tailed, and few-shot) on the CIFAR10, CIFAR100, and TinyImageNet to validate the superiority of our FedSC and the indispensability of each module.
\end{itemize}

\section{Related Work}
\subsection{Heterogeneous Federated Learning}
To mitigate the negative impact of the data heterogeneity issue in the FL training process, there are mainly two approaches: local model adjustment at the client side~\cite{li2020fedprox,gao2022federated,tan2023federated,cho2022towards,chen2023fraug} and global model adjustment at the server side~\cite{Wang2020Federated,nguyen2022federated,feddisco,3485730}. 
Most existing literature focusing on local model adjustment to improve the client training process includes, adding a regularization term to penalize client updates~\cite{li2020fedprox,yuan2022convergence}, selecting and clustering clients with similar distribution~\cite{cho2022towards,fraboni2021clustered}, using data augmentation to expand client private data against the shift~\cite{chen2023fraug,hao2021towards}, and introducing domain adaptation to align domain representations~\cite{yao2022federated}. 
More recently, federated learning combined with the knowledge distillation technique has become popular~\cite{zhu2021data,fedftg,fedbe,feddf}. 
However, these methods rely on an auxiliary dataset, which might lead to data privacy leakage in FL training. 
In another direction, a set of FL methods adopt specific mechanisms on the server to alleviate data heterogeneity. For example, FedFTG~\cite{fedftg} explores the input space of local models through a generator to fine-tune the global model on the server, FedLC~\cite{fedlc} alleviates the global drift problem under the long-tailed distribution through logit calibration. 
Although these methods could focus on aligning the local and global model to reduce drift, the gradually enlarged parameter deviation still persists. In this paper, our FedSC aims to learn a well-generalizable global model during FL training via the proposed RPCL and CPDR in a mutual-complementary manner, thereby likely getting rid of the dilemma of data heterogeneity.

\subsection{Prototype Learning}
The concept of prototypes (the representative embedding for the specific class) has been explored in a variety of downstream tasks~\cite{hou2022closer,xue2020one,wang2021interactive,xing2023boosting,lin2022text}. Formally, the class-wise prototype refers to the mean vector of the instances belonging to an identical semantic~\cite{zhou2024prototype}. For example, in the computer vision field, it labels testing images via calculating its distance with prototypes of each class, which is considered to be more robust and stable in handling few-shot~\cite{xing2023boosting}, one-shot~\cite{xue2020one}, and zero-shot~\cite{xu2020attribute} scenarios. As for the federated learning, since the prototype has the ability to generalize semantic knowledge from similar samples, it is used to assist the local training by many researchers~\cite{tan2022fedproto,mu2023fedproc,tan2023federated}. PGFL~\cite{zheng2021pose} utilizes the prototypes to construct weight attention parameter aggregation to enhance the global model. FedProto~\cite{tan2022fedproto} aims to reach feature alignment with local prototypes and global prototypes. FedProc~\cite{mu2023fedproc} proposes a global class prototype to correct for local training. FedPCL~\cite{tan2022federated} focuses on personalized federated learning and uses class-wise prototypes to learn personalized models. However, these methods only use single local prototypes while simple averaged global prototypes may suffer from poor performance. In contrast, we herein can shed light on leveraging relational prototypes and consistent prototypes to provide richer semantic knowledge and stable global signals.

\subsection{Contrastive Learning}
Contrastive learning has been widely applied to computer vision in many self-supervised learning scenarios~\cite{chen2020simple,he2020momentum,zhuang2020comprehensive,le2020contrastive}. 
A number of works focused on constructing a positive pair and a negative pair for each instance and leveraging the InfoNCE~\cite{oord2018representation} loss objective to learn an encoder. SimCLR~\cite{chen2020simple} proposed a simple framework for contrastive learning of visual representations by constructing positive and negative sample pairs via data augmentation. MoCo~\cite{he2020momentum} further built a dynamic dictionary with a queue and a moving-averaged encoder with a dictionary look-up approach to facilitating contrastive unsupervised learning. A major branch of subsequent research focused on elaborating the selection of the informative positive pairs and negative pairs~\cite{xu2024efficient,karim2022unicon,miao2022negative}. Another research direction investigated the semantic structure and introduced unsupervised clustering methods to construct representative embeddings~\cite{li2021contrastive,zhao2021graph,yang2023cluster}. There are several works that incorporate contrastive learning into federated learning to assist local training~\cite{moon2021,seo2024relaxed,tan2023federated}. 
Differently, in our work, Relational Prototypes Contrastive Learning (RPCL) is devised for promoting a more generalizable FL global model by attracting each instance to the corresponding relational prototypes in the same class while pushing away others.

\section{Federated Learning with Prototypes}
\subsection{Preliminaries} \label{sec:pre}
We first focus on the standard FL setting~\cite{li2020fedprox} with $K$ clients holding heterogeneous data partition $\{\mathcal{D}_{1}, \mathcal{D}_{2}, \ldots, \mathcal{D}_{K}\}$ with private data samples $\mathcal{D}_{k}=\left\{x_{i}, y_{i}\right\}_{i=1}^{n_{k}}, y_{i}\in\{1,\ldots,|\mathcal{C}|\}$ for the $k$-th client, where $n_{k}$ denotes the local data scale. Let $n_{k}^{j}$ be the number of samples with label $j$ at client $k$, and $\mathcal{D}_{k}^{j}=\{(x, y) \in \mathcal{D}_{k} | y=j\}$ denotes the set of samples with label $j$ on the $k$-th client, so the number of samples in $j$-th class is $n^{j}=\sum_{k=1}^{K}n_{k}^{j}$. Formally, the global objective of general FL across $K$ local clients is:
\begin{equation} \label{eq:eq1}
    \min_{w} \mathcal{L}(w) = \frac{1}{K} \sum_{k=1}^{K} \frac{n_{k}}{N} \mathcal{L}_{k} (w; \mathcal{D}_{k}),
\end{equation}
where $\mathcal{L}_{k}$ and $w$ are the local loss function and global parameters for the $k$-th client, respectively. $N$ is the total number of samples among all clients. 
In the FL training setup, we consider the local model with parameters $w=\left\{u, v\right\}$. It has two modules: a feature extractor $f(u)$ with parameters $u$ maps each sample $x_{i}$ to a $d$-dim feature vector $z_{i}=f(u;x_{i})$; a specific classifier $h(v)$ with parameters $v$ maps $z_{i}$ into a $|\mathcal{C}|$-dim output logits $l_{i}=h(v;z_{i})$. The optimization objective in Eq.~\ref{eq:eq1} is to learn a generalizable global model $w$ to present favorable performance through the FL process.

\subsection{Motivation}
The class-wise prototypes $c^{j} \in \mathbb{R}^{d}$ is computed by the mean vector of the samples' features belonging to the same label $j$:
\begin{equation} \label{eq:eq2}
    c^{j} = \frac{1}{n^{j}} \sum_{(x, y) \in \mathcal{D}^{j}} f(u; x), \; 
    \text{where} \; \mathcal{D}^{j}=\{(x, y) \in \mathcal{D} : y=j\},
\end{equation}
where $\mathcal{D}=\{\mathcal{D}_{1}, \mathcal{D}_{2}, \ldots, \mathcal{D}_{K}\}$ denotes all the data partition across $K$ clients, $n^{j}$ is the number of samples in $j$-th class across $K$ clients. The prototypes are typical for respective semantic knowledge, extracting the useful class-relevant information. However, due to the inherent data heterogeneity among the FL clients, for a single client $k$, the prototypes obtained on private data $\mathcal{D}_{k}$ tend to be biased towards local client's labeling preferences. Therefore, it motivates us to leverage the class-wise prototypes (\textit{i.e.}, semantic knowledge) from different clients to learn a generalizable FL global model without leaking sensitive privacy information. 
Next, we first define prototypes $c^{j}_{k} \in \mathbb{R}^{d}$ of the $j$-th class for the $k$-th client:
\begin{equation} \label{eq:eq3}
    c^{j}_{k} = \frac{1}{n^{j}_{k}} \sum_{(x_{i}, y_{i}) \in \mathcal{D}_{k}^{j}} f(u; x_{i}), \; 
    \mathcal{D}_{k}^{j}=\{(x_{i}, y_{i}) \in \mathcal{D}_{k} : y_{i} = j\},
\end{equation}
where $n_{k}^{j}$ is the number of data samples with label $j$ at client $k$, and $\mathcal{D}_{k}^{j}$ denotes the set of data samples with label $j$ on the $k$-th client. Then, we can further get the class-wise prototypes of all categories on client $k$, which can be formally defined as $\mathcal{P}_{k}$:
\begin{equation} \label{eq:eq4}
    \mathcal{P}_{k} = \{ c^{1}_{k}, c^{2}_{k}, \ldots, c^{|\mathcal{C}|}_{k} \} \in \mathbb{R}^{|\mathcal{C}| \times d},
\end{equation}
where $c^{j}_{k}$ are the prototypes defined in Eq.~\ref{eq:eq3} of class $j$ on client $k$, $d$ indicates the vector dimension of the prototypes. $\mathcal{P}_{k}$ is the set of prototypes for all categories $\{1, 2, \ldots, |\mathcal{C}|\}$ on the $k$-th client.

Considering the large-scale number of FL participating clients, utilizing all clients' prototypes on each class is irrational and computationally expensive. Several studies~\cite{tan2022fedproto,mu2023fedproc} came with a straightforward solution to obtain global prototypes via the directly averaging operation similar to the global aggregation in the FedAvg~\cite{McMahan17}. The global prototypes $g^{j}$ of label $j$ are computed as follows:
\begin{equation} \label{eq:eq5}
    g^{j} = \frac{1}{K} \sum_{k=1}^{K} c^{j}_{k}, \; \text{where} \; g^{j} \in \mathbb{R}^{d},
\end{equation}
where $c^{j}_{k}$ is the prototypes of the $j$-th class for the $k$-th client. Subsequently, we can further get the set $\mathcal{G}$ of global prototypes for the all categories $\{1, 2, \ldots, |\mathcal{C}|\}$ as follows:
\begin{equation} \label{eq:eq6}
    \mathcal{G} = \{ g^{1}, g^{2}, \ldots, g^{|\mathcal{C}|} \} \in \mathbb{R}^{|\mathcal{C}| \times d}.
\end{equation}
However, whether $c^{j}_{k}$ of Eq.~\ref{eq:eq3} or $g^{j}$ of Eq.~\ref{eq:eq5}, there are the following two notable problems: 
\ding{182} for the single client $k$, the prototypes $c^{j}_{k}$ learned by the local model are prone to be biased, which makes it difficult to fully generalize the semantic information with class $j$, due to the negative impact of inherent heterogeneity in the local domain; 
\ding{183} for the global prototypes $g^{j}$, it unavoidably faces the same dilemma as simply performing global averaging, because the contribution of each client is not equal, leading to a skewed optimization objective toward the dominant domain distribution in FL training. 
Inspired to solve these two issues, we further construct relational prototypes and consistent prototypes.

\begin{figure}[tbp]
    \centering
    \includegraphics[width=0.966\linewidth]{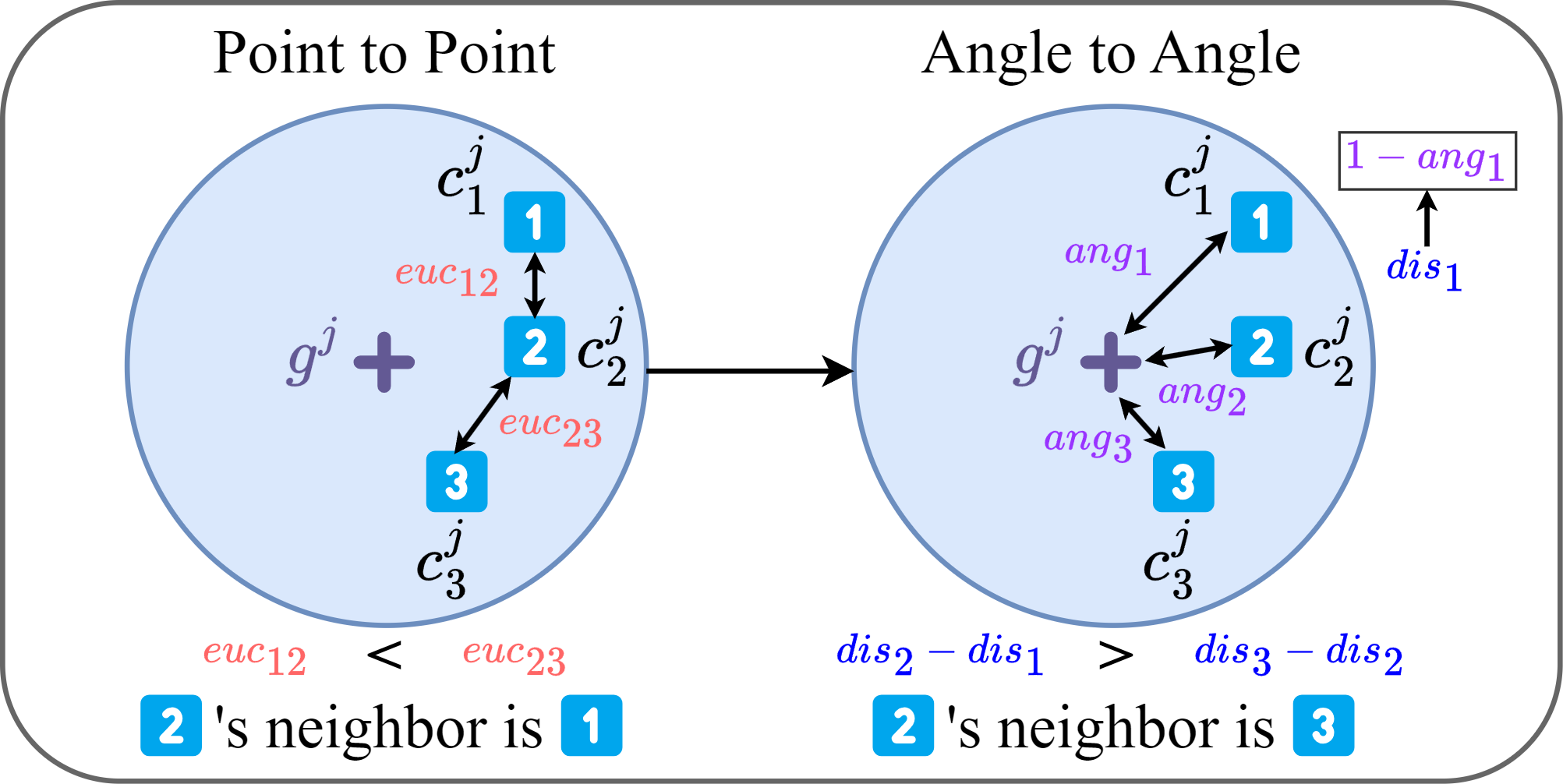}
    \caption{An example of computing neighbor relationships among prototypes by point-wise or angle-wise distance (\textcolor[RGB]{255,115,115}{$euc$} means L1 or L2 distance, \textcolor[RGB]{152,48,255}{$ang$} means cosine distance).}
    \label{fig:fig2}
\end{figure}

\begin{figure}[tbp]
    \centering
    \includegraphics[width=1.0\linewidth]{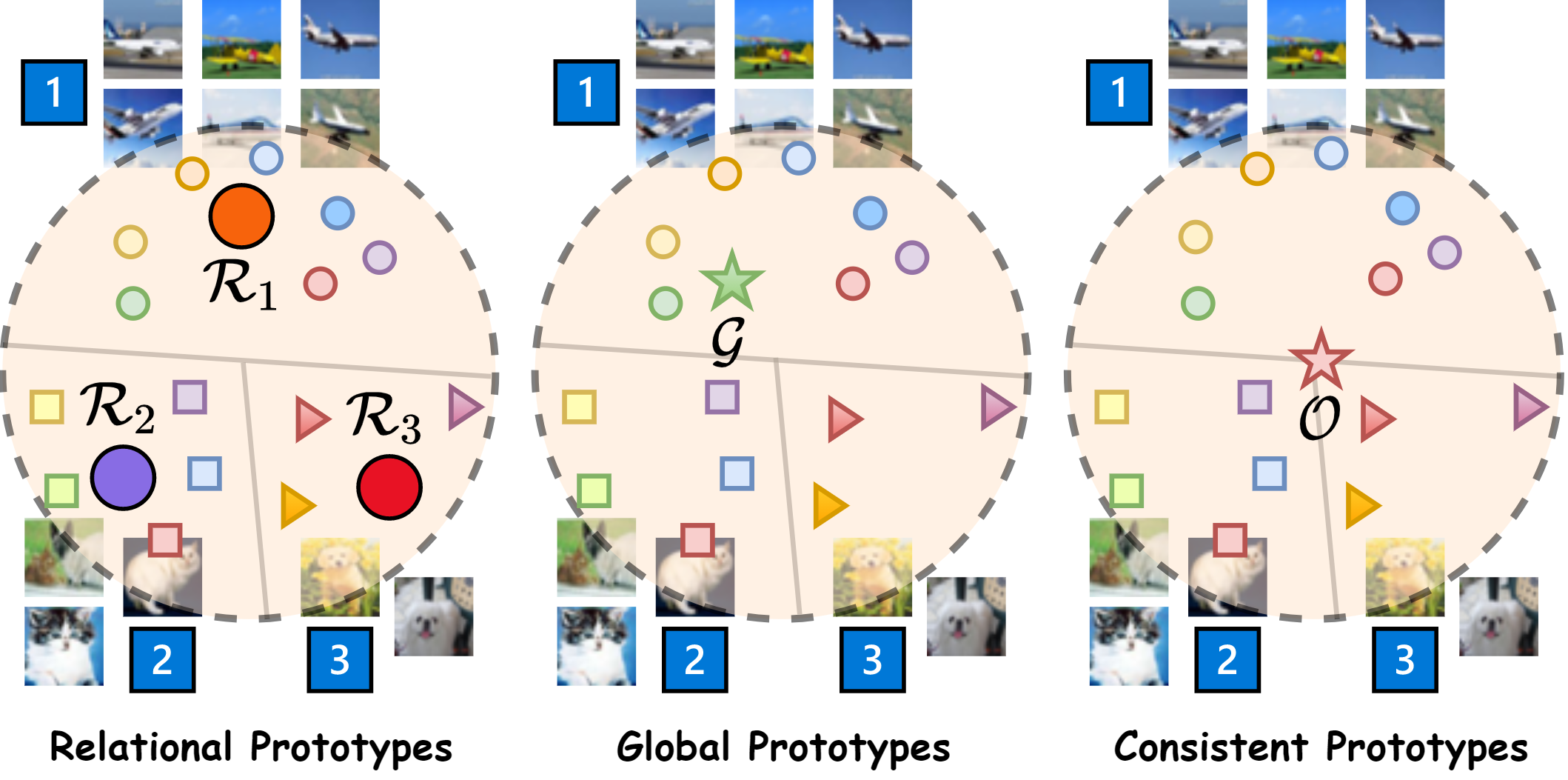}
    \caption{Illustration of different class-wise prototypes.}
    \label{fig:fig3}
\end{figure}

\begin{figure*}[tbp]
    \centering
    \includegraphics[width=0.916\linewidth]{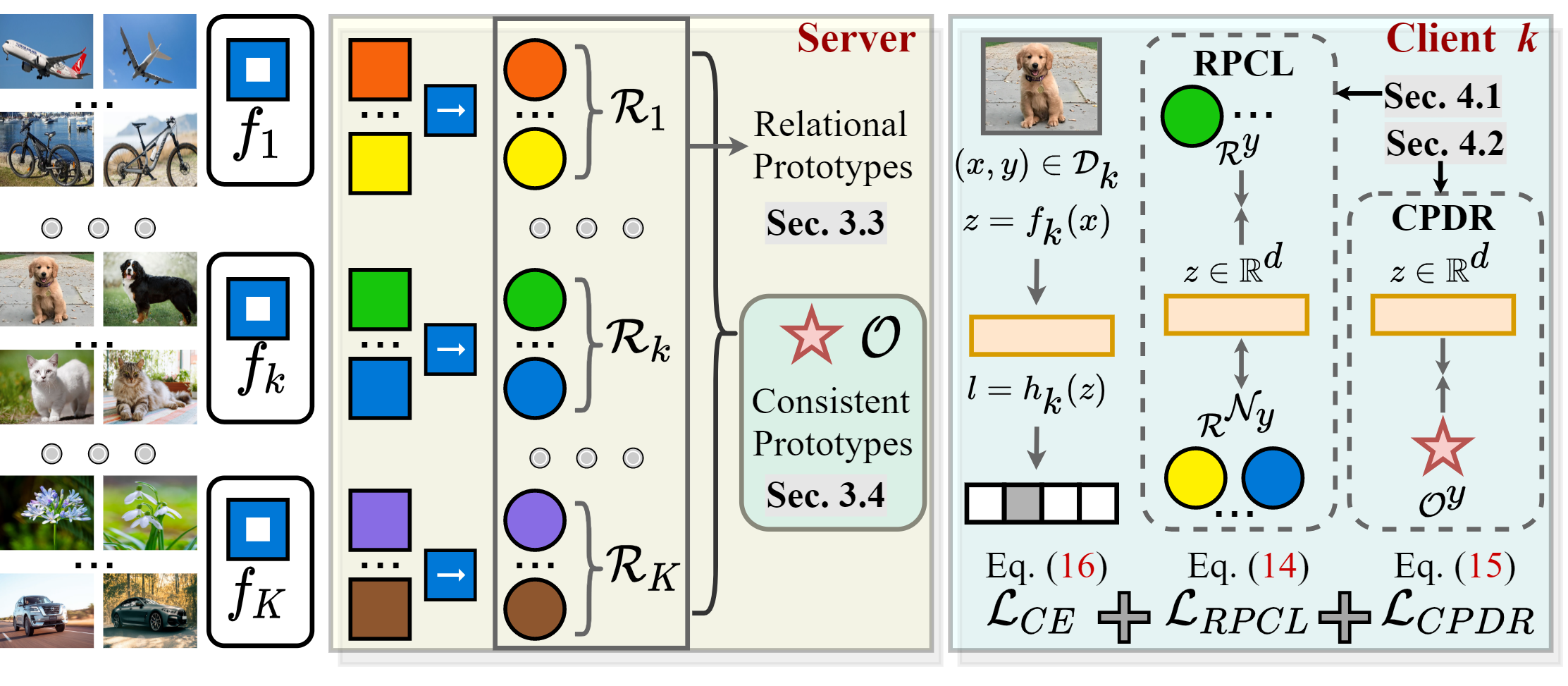}
    \caption{The overview of Federated Learning with Semantic-Aware Collaboration (FedSC), including two complementary components: RPCL (Sec.~\ref{sec:rpcl}) builds an inter-contrastive learning strategy via relational prototypes (Sec.~\ref{sec:rep}), CPDR (Sec.~\ref{sec:cpdr}) designs a regularization penalty via consistent prototypes (Sec.~\ref{sec:cop}), jointly regulating local model training.}
    \label{fig:fig4}
\end{figure*}

\subsection{Relational Prototypes} \label{sec:rep}
The key idea of the relational prototypes is based on the cognitive diversity of different local models for identical semantics, so as to merge class-wise prototypes by mutual angular relations across clients. 
Compared with the prototypes learned on a single client, we aim to devise relational prototypes in two steps: 1) we first select the semantically neighboring clients about class $j$ of the $k$-th client's prototypes $c^{j}_{k}$; 2) we then merge these prototypes that have similar representations for the same class $j$. 
Specifically, we calculate the semantic similarity among different clients' prototypes based on an angle-wise relational potential. Compared with traditional metrics methods, we capture the distance relation among prototypes by penalizing angular differences, to select semantically representative clients. 
As shown in Fig.~\ref{fig:fig2}, we assume that $g^{j}$ reflects global representations of label $j$, to a certain extent. The $c^{j}_{1}$ and $c^{j}_{2}$ are close in Euclidean space, but may not be exact neighbors due to the semantic uncertainty of the local model. 
We use the angle-wise differences to select clients with similar global semantics on prototypes. The comparison results in Table.~\ref{tab:tab5} also confirm this view. We get angular differences between all clients' prototypes and the global prototypes $g^{j}$ as follows:
\begin{equation} \label{eq:eq7}
    \phi^{j} = \left [ \frac{(g^{j})^\top c^{j}_{1}}{||g^{j}||_{2} \cdot ||c^{j}_{1}||_{2}}, \ldots, \frac{(g^{j})^\top c^{j}_{K}}{||g^{j}||_{2} \cdot ||c^{j}_{K}||_{2}} \right ] \in \mathbb{R}^{K},
\end{equation}
where $\phi^{j}$ is the set of all angular differences for class $j$, we define $\phi^{j}_{k}$ is the $k$-th client's angular difference. Then, based on $\phi^{j}$ in Eq.~\ref{eq:eq7}, we select top $M$ neighbor clients closest to $\phi^{j}_{k}$. 
We further build an adjacency matrix $\mathcal{A}^{j}$ that summarizes similar prototypes of class $j$ into the same group, which can be formulated as:
\begin{equation} \label{eq:eq8}
    \mathcal{A}^{j}(k_{1}, k_{2})=\left\{
    \begin{aligned}
    1 & , \; \text{if} \enspace k_{2} \in \text{argsort}^{M}_{k} |\phi^{j}_{k_{1}}-\phi^{j}_{k}| 
    \enspace \text{or} \enspace k_{2}=k_{1}; \\
    0 & , \; \text{otherwise}.
    \end{aligned}\right.
\end{equation}
where $\text{argsort}^{M}_{k}$ denotes the top-$M$ indices that would sort an array from small to large, 
$k\in\{1,\ldots,K\}\setminus\{k_{1}\}$, and $k_{1},k_{2}\in\{1,\ldots,K\}$. 
Based on Eq.~\ref{eq:eq8}, we select similar clients of $k$-th client and merge these clients' prototypes to obtain relational prototypes:
\begin{equation} \label{eq:eq9}
\begin{aligned}
    r^{j}_{k} = \frac{1}{M+1} \sum_{q=1}^{K} \mathcal{A}^{j}(k, q) \cdot c^{j}_{q} \in \mathbb{R}^{d}, \\
    \mathcal{R}_{k} = \{ r^{1}_{k}, r^{2}_{k}, \ldots, r^{|\mathcal{C}|}_{k} \} \in \mathbb{R}^{|\mathcal{C}| \times d},
\end{aligned}
\end{equation}
where $r^{j}_{k}$ is $k$-th client's relational prototypes for class $j$, $\mathcal{R}_{k}$ is the set of relational prototypes for all categories $\{1,2,\ldots,|\mathcal{C}|\}$ on client $k$. 
We combine the semantically neighboring clients' prototypes to generate relational prototypes to more richly represent the latent semantics information, which effectively addresses the problem \ding{182}.

\subsection{Consistent Prototypes} \label{sec:cop}
Although we devise a corresponding relational prototypes $\mathcal{R}_{k}$ for each client $k$, $\mathcal{R}_{k}$ is still highly variable and depends on the local model, which can not ensure a flat and stable convergent point. 
Therefore, we further design consistent prototypes to capture global class-relevant semantic information. We argue that simply averaging relational prototypes across all clients is not suitable, because the contribution of each client's relational prototypes is not consistent. For each client, we compute the distance between its local category distribution and the hypothetical global category distribution, treating as discrepancy weights to integrate relational prototypes. We consider that the hypothetical global category distribution is uniform, which promotes the fairness of the semantic information across all classes. Concretely, we set the hypothetical global category distribution as a vector $\mathcal{H}=[\frac{1}{|\mathcal{C}|},\ldots,\frac{1}{|\mathcal{C}|}] \in \mathbb{R}^{|\mathcal{C}|}$, the value of each element in $\mathcal{H}$ is $1/|\mathcal{C}|$. 
For the client $k$, the local category distribution $\mathcal{B}_{k}=[\frac{n^{1}_{k}}{n_{k}}, \ldots, \frac{n^{|\mathcal{C}|}_{k}}{n_{k}}] \in \mathbb{R}^{|\mathcal{C}|}$ is abstracted as a vector, where $n_{k}^{j}$ is the number of samples with $j$-th class at client $k$ and $n_{k}$ is the total number of samples at client $k$. Then, the $k$-th client's discrepancy $d_{k}$ is formulated as:
\begin{equation} \label{eq:eq10}
    d_{k}=\sqrt{\frac{1}{2} \sum_{j=1}^{|\mathcal{C}|} \left(\mathcal{B}^{j}_{k} - \mathcal{H}^{j}\right)^{\mathbf{2}}}, 
    \; \text{where} \; \mathcal{B}^{j}_{k}=\frac{n^{j}_{k}}{n_{k}}, \; \mathcal{H}^{j}=\frac{1}{|\mathcal{C}|},
\end{equation}
where $\mathcal{B}^{j}_{k}$ is the value of $\mathcal{B}_{k}$ on class $j$ and $\mathcal{H}^{j}$ is the value of $\mathcal{H}$ on class $j$. Based on Eq.~\ref{eq:eq10}, we get the discrepancy value for each client. 
We can speculate that the client $k$ with a larger sample size $n_{k}$ and smaller discrepancy $d_{k}$ would be assigned a higher aggregation weight $e_{k}$ when integrating relational prototypes:
\begin{equation} \label{eq:eq11}
    e_{k} = \frac{\text{Sigmoid}(a \cdot n_{k} - b \cdot d_{k})} 
    {\sum_{i=1}^{K}\text{Sigmoid}(a \cdot n_{i} - b \cdot d_{i})},
\end{equation}
where $a=1/N$ and $b=1 / \sum_{i=1}^{K}d_{i}$ are used to balance the $n_{k}$ and $d_{k}$, $n_{k}$ is the total number of the $k$-th client's samples, $N$ is the total number of samples of all clients. 
Based on Eq.~\ref{eq:eq11}, even if the client sample size is large (\textit{e.g.}, long-tailed scenario), we can determine a more distinguishing weight for each client rather than a simple average. We then further obtain consistent prototypes $\mathcal{O}$ as follows:
\begin{equation} \label{eq:eq12}
\begin{aligned}
    & o^{j} = \sum_{k=1}^{K} e_{k} \cdot r^{j}_{k} \in \mathbb{R}^{d}, \\
    & \mathcal{O} = \{ o^{1}, o^{2}, \ldots, o^{|\mathcal{C}|} \} \in \mathbb{R}^{|\mathcal{C}| \times d},
\end{aligned}
\end{equation}
where $r^{j}_{k}$ is $k$-th client's relational prototypes for class $j$ in Eq.~\ref{eq:eq9}. Thus, we are able to hypothesize that consistent prototypes $\mathcal{O}$ could depict the considerably flat and stable convergent point and further leverage them as a regularization penalty to constrain the optimization region of the local model, which would effectively handle the problem \ding{183}. We illustrate and explain the difference of these three kinds of prototypes (relational prototypes $\mathcal{R}_{k}$, global prototypes $\mathcal{G}$, and consistent prototypes $\mathcal{O}$) in Fig.~\ref{fig:fig3}. Notably, both relational and consistent prototypes are privacy-preserving because they are yielded from multiple averaging operations~\cite{zhu2021prototype,tan2022fedproto}, and avoid the exposure of sensitive information compared to the original model features. Therefore, leveraging $\mathcal{R}_{k}$ and $\mathcal{O}$, is not only a computation-friendly method but also a privacy-preserving solution.

\section{FL with Semantic-Aware Collaboration}
We leverage relational prototypes and consistent prototypes to obtain rich class-relevant domain diversity and flat stable global signals by two complementary modules: Relational Prototypes Contrastive Learning (RPCL) in Sec.~\ref{sec:rpcl} and Consistent Prototypes Discrepancy Regularization (CPDR) in Sec.~\ref{sec:cpdr}. Finally, we provide a theoretical analysis of FedSC under non-convex objectives in Sec.~\ref{sec:ana} to ensure convergence and a guaranteed convergence rate. 
The overview of our proposed FedSC is shown in Fig.~\ref{fig:fig4}.

\subsection{Relational Contrastive Learning} \label{sec:rpcl}
Inspired by the success of the contrastive learning~\cite{chen2020simple} and transfer learning~\cite{zhuang2020comprehensive}, researchers have argued that a well-generalizable local model should provide clear decision boundaries for different labels instead of biased boundaries limited to the local dataset. In our method, further, for the label $j$, the relational prototypes $\mathcal{R}^{j}=\{r^{j}_{1}, r^{j}_{2}, \ldots, r^{j}_{K}\} \in \mathbb{R}^{K \times d}$ from different clients can be regarded as virtual teachers for input samples on the client $k$. In other words, for the dataset $\mathcal{D}_{k}=\{x_{i}, y_{i}\}_{i=1}^{n_{k}}$ on $k$-th client, each sample's feature $z_{i}=f_{k}(x_{i}|y_{i}=j) \in \mathbb{R}^{d}$ and $\mathcal{R}^{j} \in \mathbb{R}^{K \times d}$ constitute positive sample pairs, and the feature vector $z_{i} \in \mathbb{R}^{d}$ and $\mathcal{R}^{\mathcal{N}_{j}} \in \mathbb{R}^{(|\mathcal{C}|-1) \times K \times d}$ (\textit{i.e.}, $\mathcal{R}^{\mathcal{N}_{j}}=\mathcal{R}-\mathcal{R}^{j}$) constitute negative sample pairs. Therefore, we further propose Relational Prototypes Contrastive Learning (RPCL) to maximize the similarity between positive pairs ($z_{i}$ and $\mathcal{R}^{j}$) while minimizing the similarity between negative pairs ($z_{i}$ and $\mathcal{R}^{\mathcal{N}_{j}}$). We define the similarity $s(z_{i}, r^{j}_{k})$ of the sample output feature $z_{i}$ with relational prototypes $r^{j}_{k} \in \mathcal{R}^{j}$ as:
\begin{equation} \label{eq:eq13}
    s(z_{i}, r^{j}_{k})=\frac{1}{\mathcal{U}}(\frac{z_{i} \cdot r^{j}_{k}}{\|z_{i}\|_{2} \times \|r^{j}_{k}\|_{2}}), \; \mathcal{U}=\frac{1}{n_{k}}\sum_{q=1}^{n_{k}}\left\|z_{q}-r^{j}_{k}\right\|_{2},
\end{equation}
where $n_{k}$ is total number of samples at client $k$, $\mathcal{U}$ is a normalization factor for similarity, where we set $\mathcal{U}$ to denote the average distance between $z_{i}|_{i=1}^{n_{k}}$ and $r^{j}_{k}$ on dataset $\mathcal{D}_{k}$. In this way, based on Eq.~\ref{eq:eq13}, we can measure the similarity among different sample pairs. Then, we introduce RPCL to quantify and regulate local model training, and it can be formulated as the following optimization objective:
\begin{equation} \label{eq:eq14}
    \begin{aligned}
    \mathcal{L}_{RPCL} & =-\text{log}\frac{\sum_{r^{j} \in \mathcal{R}^{j}} \text{exp}(s(z_{i},r^{j})/\tau)}{\sum_{r^{j} \in \mathcal{R}^{j}} \text{exp}(s(z_{i},r^{j})/\tau) + \sum_{r^{q} \in \mathcal{R}^{\mathcal{N}_{j}}} \text{exp}(s(z_{i},r^{q})/\tau)} \\
    & \equiv \mathbf{\min} (\text{log} \frac{\sum_{r^{q} \in \mathcal{R}^{\mathcal{N}_{j}}} \text{exp}(s(z_{i},r^{q})/\tau) \mathcolorbox{myblue}{\Rightarrow \mathbf{Exclusiveness}}}{\sum_{r^{j} \in \mathcal{R}^{j}} \text{exp}(s(z_{i}, r^{j})/\tau) \mathcolorbox{mygreen}{\Rightarrow \mathbf{Attractiveness}}}),
    \end{aligned}
\end{equation}
where $\tau$ denotes a temperature parameter to control the representation strength~\cite{chen2020simple}. For client $k$, note that by minimizing Eq.~\ref{eq:eq14}, the local model brings each sample output feature $z_{i}$ closer to relational prototypes $\mathcal{R}^{j}$ (\mathcolorbox{mygreen}{\Rightarrow \mathbf{Attractiveness}}) and away from relational prototypes $\mathcal{R}^{\mathcal{N}_{j}}$ of other classes (\mathcolorbox{myblue}{\Rightarrow \mathbf{Exclusiveness}}). In other words, relational prototypes $\mathcal{R}^{j}$ can be regarded as a virtual teacher to regulate the local model training on the client's private dataset to maintain a clear class-wise decision boundary for class $j$, promoting a satisfactory generalizable performance in FL.

\subsection{Consistent Discrepancy Regularization} \label{sec:cpdr}
Although the relational prototypes provide diverse class-relevant semantic knowledge across clients by minimizing Eq.~\ref{eq:eq14}, relational prototypes are dynamically generated at each round and its scale might be changing due to the local private dataset. To force the sample feature to be closer to its corresponding global class centroid so as to extract more class-relevant, but client-irrelevant information, we define consistent prototypes $\mathcal{O}$ based on relational prototypes $\mathcal{R}$ in Eq.~\ref{eq:eq12} via a discrepancy aggregation manner. 
We assume that consistent prototypes $\mathcal{O}$ could offer a relatively flat and stable optimization point at different communication rounds and thus cope with the problem of convergence instability. 
Intuitively, the consistent prototypes $\mathcal{O}$ consider the discrepancy between local category distribution and global category distribution, providing an opportunity to learn from a summarized client-irrelevant perspective. Concretely, we leverage consistent prototypes $\mathcal{O}$ to constrain the optimization region of the local model, and propose Consistent Prototypes Discrepancy Regularization (CPDR). 
Given an input sample $\{x_{i}, y_{i}=j\}$, we use a regularization penalty to pull the sample feature $z_{i}=f(x_{i}|y_{i}=j)$ closer to the respective consistent prototypes $\mathcal{O}^{j}=o^{j}$ in Eq.~\ref{eq:eq12}, which is formulated as follows:
\begin{equation} \label{eq:eq15}
    \mathcal{L}_{CPDR} = \sum_{q=1}^{d} \left\| z_{i(q)}-\mathcal{O}_{(q)}^{j} \right\|_{2}, \; z_{i} \in \mathbb{R}^{d}, \; \mathcal{O}^{j} \in \mathbb{R}^{d},
\end{equation}
where $q$ indexes the dimension $d$ of feature output. Based on Eq.~\ref{eq:eq15}, we expect to align the instance embedding with the corresponding consistent prototype in the feature level.

Besides, we construct a CrossEntropy loss~\cite{celoss2005} and use the logits output vector $l_{i}=h(z_{i})$ with the original label $y_{i}$ to optimize local discriminative ability for each client as follows:
\begin{equation} \label{eq:eq16}
    \mathcal{L}_{CE} = -\mathbf{1}_{y_{i}} \log (\psi(h(z_{i}))), \; z_{i} \in \mathbb{R}^{d},
\end{equation}
where $\psi$ denotes the Softmax function, $h$ is the classifier with the parameters $v$ maps $z_{i} \in \mathbb{R}^{d}$ into a $|\mathcal{C}|$-dim output logits $l_{i}$. Finally, for the $k$-th client, we carry out the following optimization objective $\mathcal{L}=\mathcal{L}_{RPCL}(\text{Eq.}~\ref{eq:eq14})+\mathcal{L}_{CPDR}(\text{Eq.}~\ref{eq:eq15})+\mathcal{L}_{CE}(\text{Eq.}~\ref{eq:eq16})$ in the local training. 
The pseudo-code of FedSC is provided in Algorithm~\ref{alg:fedsc}.

\renewcommand{\algorithmicrequire}{\textbf{RunServer}($w_{k}, \mathcal{P}_{k}$):}  
\renewcommand{\algorithmicensure}{\textbf{RunClient}($w_{g}, \mathcal{R}, \mathcal{O}$):} 
\begin{algorithm}[htbp]
    \caption{Pseudo-code of FedSC.}
    \label{alg:fedsc}
    \begin{algorithmic}[1]
        \Require \emph{\textcolor[RGB]{0,0,255}{/* \quad Global Server \quad */}} 
        \State Initialized global rounds $R$, number of clients $K$ and labels $|\mathcal{C}|$, 
                client $k$ dataset $\mathcal{D}_{k}(x, y)$, client model $w_{k}$, global model $w_{g}$
        \State Note: Clients' Prototypes $\mathcal{P}$ via Eq.~\ref{eq:eq4}, Relational Prototypes $\mathcal{R}$ via Eq.~\ref{eq:eq9}, Consistent Prototypes $\mathcal{O}$ via Eq.~\ref{eq:eq12}
        \For{each round $r = 1,2,...,R$}
            \State $A_{r} \leftarrow$ (server selects a random subset of $K$ clients)
            \For{each client $ k \in A_{r} $ \textbf{in parallel}}
                \State $w_{k}, \mathcal{P}_{k} \leftarrow \text{RunClient}(w_{g}, \mathcal{R}, \mathcal{O})$
            \EndFor
            \State $w_{g} \leftarrow \sum_{k=1}^{|A_{r}|} \frac{n_{k}}{N} w_{k}$ \emph{\textcolor[RGB]{0,0,255}{// global model}}
            \State \emph{\textcolor[RGB]{0,0,255}{// Relational Prototypes (Sec.~\ref{sec:rep})}}
            \State $\phi \leftarrow$ (angular differences via Eq.~\ref{eq:eq7})
            \State $\mathcal{A} \leftarrow$ (adjacency matrix via Eq.~\ref{eq:eq8})
            \State $\mathcal{R} \leftarrow \{\mathcal{P},\phi,\mathcal{A}\}$ (relational prototypes via Eq.~\ref{eq:eq9})
            \State \emph{\textcolor[RGB]{0,0,255}{// Consistent Prototypes (Sec.~\ref{sec:cop})}}
            \State $d \leftarrow$ (clients' discrepancy via Eq.~\ref{eq:eq10})
            \State $e \leftarrow$ (aggregation weights via Eq.~\ref{eq:eq11})
            \State $\mathcal{O} \leftarrow \{\mathcal{R},d,e\}$ (consistent prototypes via Eq.~\ref{eq:eq12})
        \EndFor
        \Ensure \emph{\textcolor[RGB]{0,0,255}{/* \quad Local Client \quad */}} 
        \State Initialized local epochs $E$, size of samples $n_{k}$, learning rate $\eta$
        \For{each local epoch $\mathbf{e} = 1,2,...,E$} 
            \For{each batch $b \in$ client dataset $\mathcal{D}_{k}$}
                \State $\mathcal{L}_{CE}=\sum_{i\in b} -\mathbf{1}_{y_{i}} \log (\psi(h(z_{i})))$ 
                    \emph{\textcolor[RGB]{0,0,255}{// via Eq.~\ref{eq:eq16}}}
                \State \emph{\textcolor[RGB]{0,0,255}{// Relational Prototypes Contrastive Learning}}
                \State $\mathcal{L}_{RPCL} \leftarrow \{z_{i}, \mathcal{R}\}$ (Eq.~\ref{eq:eq14} in Sec.~\ref{sec:rpcl})
                \State \emph{\textcolor[RGB]{0,0,255}{// Consistent Prototypes Discrepancy Regularization}}
                \State $\mathcal{L}_{CPDR} \leftarrow \{z_{i}, \mathcal{O}\}$ (Eq.~\ref{eq:eq15} in Sec.~\ref{sec:cpdr})
                \State $\mathcal{L} = \mathcal{L}_{CE} + \mathcal{L}_{RPCL} + \mathcal{L}_{CPDR}$
                \State $w_{k} \leftarrow w_{k}-\eta \nabla \mathcal{L}(w_{k}; b)$ \emph{\textcolor[RGB]{0,0,255}{// model update by SGD}}
            \EndFor	
        \EndFor
        \State \emph{\textcolor[RGB]{0,0,255}{// Client's Prototypes Calculation}}
        \For{$k=1,2,...,n_{k}$}
            \State $\mathcal{D}_{k}^{j}=\{(x_{i}, y_{i}) \in \mathcal{D}_{k} | y_{i}=j \} \subset \mathcal{D}_{k}$
            \State $c^{j}_{k} = \frac{1}{n^{j}_{k}} \sum_{(x_{i}, y_{i}) \in \mathcal{D}_{k}^{j}} f(x_{i})$ 
                \emph{\textcolor[RGB]{0,0,255}{// via Eq.~\ref{eq:eq3}}}
        \EndFor
        \State $\mathcal{P}_{k} = \{c^{1}_{k}, \ldots, c^{|\mathcal{C}|}_{k}\}$ \emph{\textcolor[RGB]{0,0,255}{// via Eq.~\ref{eq:eq4}}}
        \State \textbf{return} $w_{k}$, $\mathcal{P}_{k}$
    \end{algorithmic}
\end{algorithm}

\begin{table*}[t]
\caption{Accuracy comparisons on several heterogeneous settings and datasets, $\mathtt{NID1}_{\alpha}$ means Dirichlet distribution with ratio $\alpha$. Best in bold and second with underline. FedSC consistently outperforms these state-of-the-art methods.}
\label{tab:tab1}
\centering
\resizebox{\linewidth}{!}{ 
\begin{tabular}{l||ccc|ccc|ccc|ccc}
    \Xhline{1.2pt}
    \rowcolor{mygray} & \multicolumn{3}{c}{\textbf{CIFAR10}} & \multicolumn{3}{c}{\textbf{CIFAR100}} & \multicolumn{3}{c}{\textbf{TinyImageNet}} & \multicolumn{3}{c}{\textbf{FC100} (5-way 5-shot)} \\
    \cline{2-13}
    \rowcolor{mygray} \multirow{-1.8}{*}{\textbf{Methods}} & $\mathtt{NID1}_{0.05}$ & $\mathtt{NID1}_{0.2}$ & $\mathtt{NID2}$ & $\mathtt{NID1}_{0.05}$ & $\mathtt{NID1}_{0.2}$ & $\mathtt{NID2}$ & $\mathtt{NID1}_{0.05}$ & $\mathtt{NID1}_{0.2}$ & $\mathtt{NID2}$ & $\mathtt{NID1}_{0.05}$ & $\mathtt{NID1}_{0.2}$ & $\mathtt{NID2}$ \\
    \hline\hline
    $\text{FedAvg}_{[\text{AISTAT'17}]}$ & 78.27 & 84.65 & 75.60 & 55.97 & 60.08 & 54.56 & 40.41 & 42.84 & 37.15 & 47.40 & 50.24 & 45.21\\
    $\text{FedProx}_{[\text{MLSys'20}]}$ & 79.33 & 84.59 & 76.01 & 56.27 & 60.21 & 54.29 & 40.20 & 42.16 & 37.66 & 47.06 & 50.43 & 45.86\\
    $\text{MOON}_{[\text{CVPR'21}]}$ & 80.23 & 86.10 & 77.35 & 56.79 & 61.48 & 55.81 & 40.23 & 44.03 & 38.25 & 47.82 & 51.12 & 46.17\\
    $\text{FedNTD}_{[\text{NIPS'22}]}$ & 80.34 & 86.23 & 76.75 & 57.12 & 61.74 & 56.33 & 41.19 & 43.92 & 38.37 & 48.23 & 51.22 & 47.14\\
    $\text{FedProto}_{[\text{AAAI'22}]}$ & 76.86 & 83.90 & 73.84 & 52.48 & 58.26 & 52.60 & 38.29 & 41.02 & 35.86 & 45.31 & 47.23 & 43.37\\
    $\text{FedNH}_{[\text{AAAI'23}]}$ & 79.34 & 85.25 & 75.66 & 57.64 & 60.43 & 55.70 & 40.63 & 43.20 & 38.26 & \underline{48.54} & 50.82 & 46.06\\
    $\text{FedRCL}_{[\text{CVPR'24}]}$ & \underline{81.30} & 86.89 & \underline{78.32} & \underline{58.35} & \underline{62.36} & 56.51 & 41.48 & \underline{44.34} & 38.81 & 48.13 & \underline{51.35} & \underline{47.21}\\
    $\text{FedCDA}_{[\text{ICLR'24}]}$ & 80.79 & \underline{87.11} & 77.81 & 58.12 & 61.95 & \underline{56.62} & \underline{41.56} & 44.27 & \underline{38.92} & 47.95 & 51.10 & 46.97\\
    \hline\hline
    \rowcolor{myye} FedSC & \textbf{82.41} & \textbf{88.13} & \textbf{79.61} & \textbf{59.18} & \textbf{63.25} & \textbf{57.83} & \textbf{42.63} & \textbf{45.10} & \textbf{40.09} & \textbf{49.62} & \textbf{52.28} & \textbf{48.03} \\
    \Xhline{1.2pt}
\end{tabular}}
\end{table*}

\begin{table*}[t]
\caption{Accuracy comparisons on several unbalanced settings and datasets, $\rho$ means the ratio between sample sizes of the most frequent and least frequent class, \textit{i.e.}, $\rho=\max_{j}\{n_{j}\} / \min_{j}\{n_{j}\}$, and we set these datasets under the same setting $\mathtt{NID1}_{0.2}$.}
\label{tab:tab2}
\centering
\resizebox{\linewidth}{!}{ 
\begin{tabular}{l||cccc|cccc|cccc}
    \Xhline{1.2px}
    \rowcolor{mygray} & \multicolumn{4}{c}{\textbf{CIFAR10-LT} ($\mathtt{NID1}_{0.2}$)} & \multicolumn{4}{c}{\textbf{CIFAR100-LT} ($\mathtt{NID1}_{0.2}$)} & \multicolumn{4}{c}{\textbf{TinyImageNet-LT} ($\mathtt{NID1}_{0.2}$)} \\
    \cline{2-13}
    \rowcolor{mygray} \multirow{-1.8}{*}{\textbf{Methods}} & $\rho=10$ & $\rho=50$ & $\rho=100$ & $\rho=200$ & $\rho=10$ & $\rho=50$ & $\rho=100$ & $\rho=200$ & $\rho=10$ & $\rho=50$ & $\rho=100$ & $\rho=200$ \\
    \hline\hline
    $\text{FedAvg}_{[\text{AISTAT'17}]}$ & 75.94 & 63.25 & 52.97 & 49.69 & 45.88 & 35.31 & 32.48 & 28.57 & 28.80 & 23.87 & 21.33 & 19.49\\
    $\text{FedProx}_{[\text{MLSys'20}]}$ & 75.17 & 63.09 & 52.45 & 50.03 & 45.49 & 36.25 & 31.67 & 29.12 & 29.21 & 23.66 & 20.82 & 20.25\\
    $\text{MOON}_{[\text{CVPR'21}]}$ & 76.02 & 63.65 & 53.10 & 49.93 & 46.15 & 37.81 & 33.43 & 30.75 & 30.95 & 24.89 & 22.19 & 20.68\\
    $\text{FedNTD}_{[\text{NIPS'22}]}$ & 76.10 & 63.87 & 52.79 & 50.15 & 46.52 & 38.07 & 33.64 & 31.25 & 30.17 & 24.27 & 22.58 & 21.45\\
    $\text{FedProto}_{[\text{AAAI'22}]}$ & 72.49 & 58.77 & 49.54 & 45.82 & 42.26 & 34.20 & 31.80 & 25.79 & 25.35 & 20.54 & 19.65 & 16.23\\
    $\text{FedNH}_{[\text{AAAI'23}]}$ & 75.31 & 62.50 & 52.38 & 49.17 & 45.59 & 35.39 & 32.35 & 29.08 & 28.18 & 23.93 & 21.19 & 20.14\\
    $\text{FedRCL}_{[\text{CVPR'24}]}$ & \underline{76.45} & 64.11 & 53.92 & 50.08 & 47.37 & \underline{38.70} & 34.95 & 32.76 & \underline{31.43} & 25.26 & 23.31 & 22.16\\
    $\text{FedCDA}_{[\text{ICLR'24}]}$ & 76.26 & \underline{64.58} & \underline{54.49} & \underline{50.33} & \underline{47.56} & 38.51 & \underline{35.28} & \underline{33.62} & 31.22 & \underline{25.53} & \underline{23.42} & \underline{22.91}\\
    \hline\hline
    \rowcolor{myye} FedSC & \textbf{77.31} & \textbf{65.30} & \textbf{56.34} & \textbf{51.12} & \textbf{49.10} & \textbf{40.16} & \textbf{36.52} & \textbf{34.93} & \textbf{32.57} & \textbf{26.22} & \textbf{25.15} & \textbf{24.78} \\
    \Xhline{1.2px}
\end{tabular}}
\end{table*}

\subsection{Convergence Analysis of FedSC} \label{sec:ana}
We herein further provide insights into the convergence analysis for our FedSC. 
We denote the objective function $\mathcal{L}=\mathcal{L}_{RPCL}+\mathcal{L}_{CPDR}+\mathcal{L}_{CE}$ as $\mathcal{L}_{r}$ with a subscript $r$ indicating the number of rounds, and we make similar assumptions as the existing FL research works, such as~\cite{tan2022fedproto,li2020fedprox,feddisco,Li2020On}.

Based on the assumptions (details in Appendix~\ref{app:c2}), we present the theoretical results under non-convex objectives. 
We express $\mathbf{e} \in \{1,2,\ldots,E\}$ as the local iteration step, $r$ as the global round, $rE+\mathbf{e}$ refers to the $\mathbf{e}$-th local update in round $r+1$, $rE+1$ denotes the time between $r$-th global aggregation at the server and starting the local model update on round $r+1$ at the client.
\begin{theorem}[Deviation bound of the objective function] \label{theorem1}
    Under the assumptions in Appendix~\ref{app:c2}, after every communication round, the objective function $\mathcal{L}$ of an arbitrary client will be bounded,
    \begin{equation}
        \begin{aligned}
        \mathbb{E}[\mathcal{L}_{(r+1)E}] & \leq \mathcal{L}_{rE}-\left(\eta-\frac{L_{1}\eta^{2}}{2}\right)EB^{2}+
        \frac{L_{1}E\eta^{2}}{2}\sigma^{2} \\
        & +\frac{L_{2}E\eta|\mathcal{C}|B(M+2)}{M+1},
        \end{aligned}
    \end{equation}
\end{theorem}
where $\eta$ is the learning rate, $\{|\mathcal{C}|,M\}$ are constants from Sec.~\ref{sec:rep}, $\{L_{1},L_{2},B,\sigma^{2}\}$ are constants in the assumptions from Appendix~\ref{app:c2}. Theorem~\ref{theorem1} indicates that $\mathcal{L}_{r}$ decreases as round $r$ increases, and the convergence can be guaranteed by choosing an appropriate $\eta$.
\begin{theorem}[Non-convex convergence of FedSC] \label{theorem2}
    Under the assumptions, the objective function $\mathcal{L}$ of an arbitrary client decreases monotonically as the round increases with the following condition,
    \begin{equation}
        \begin{aligned}
        \eta_{\mathbf{e}} < \frac{2(M+1)B^{2}-2(M+2)L_{2}|\mathcal{C}|B}{L_{1}(M+1)(\sigma^{2}+B^{2})}, \enspace \mathbf{e} \in \{1,2,\ldots,E-1\}.
        \end{aligned}
    \end{equation}
\end{theorem}
\begin{theorem}[Non-convex convergence rate of FedSC] \label{theorem3}
    Let the assumptions hold, the round $r$ from $0$ to $R-1$, given any $\xi>0$, the objective function $\mathcal{L}$ will converge when,
    \begin{equation}
        \begin{aligned}
        R > \frac{2(M+1)(\mathcal{L}_{0}-\mathcal{L}^{*})}{\xi E\eta(M+1)(2-L_{1}\eta)- (\varOmega_{1} + \varOmega_{2})},
        \end{aligned}
    \end{equation}
    and further get the following condition for $\eta$,
    \begin{equation}
    \eta < \frac{2 \xi (M+1) - 2(M+2)L_{2}|\mathcal{C}|B}
            {L_{1}(M+1)(\xi + \sigma^{2})},
    \end{equation}
\end{theorem}
where $\varOmega_{1}=(M+1)L_{1}E\eta^{2}\sigma^{2}$, $\varOmega_{2}=2(M+2)L_{2}E\eta|\mathcal{C}|B$, and $\mathcal{L}^{*}$ denotes the optimal solution. Theorem~\ref{theorem3} provides the convergence rate of FedSC, the smaller $\xi$ is, the larger $R$ is, which means that the tighter the bound is, more communication rounds $R$ is required. \textit{The all assumptions and proofs are formally given in Appendix~\ref{app:c}}.

\section{Experiments}
\subsection{Experimental Setup}
\textbf{Datasets:} We adopt four popular datasets, including CIFAR10 \& CIFAR100~\cite{cifar2009krizhevsky}, TinyImageNet~\cite{le2015tiny} for image classification; FC100~\cite{fc1002018tadam}, a split dataset based on CIFAR100 for few-shot classification. \\
\textbf{Scenarios:} 1) Label shift: we consider two \textit{non-iid} settings. $\mathtt{NID1}$ follows Dirichlet distribution~\cite{Wang2020Federated}, where $\alpha$ (default 0.2) is an argument correlated with heterogeneity level, and we set 10 clients; $\mathtt{NID2}$ is a more heterogeneous setting consists of 5 biased clients (each has $|\mathcal{C}|/5$ classes) and 1 client has all classes. 2) Imbalance shift: we shape the original dataset into a long-tailed distribution with $\rho$ follow~\cite{caoKaidi2019}, \textit{e.g.}, CIFAR10-LT is the unbalanced version (number of samples per class is different). 3) Few-shot shift: this setting uses the FC100 dataset and the classification is based on the Euclidean distances between query samples and support samples.

\noindent \textbf{Configurations:} Unless otherwise mentioned, we conduct $R=100$ global rounds with $K=10$ clients and local updating epochs $E=10$, where all methods have little or no accuracy gain with more rounds. To facilitate a fair comparison, we follow the setting in~\cite{li2020fedprox,moon2021}, we use SGD optimizer to update models with a learning rate of $0.01$, a momentum of $0.9$, a batch size of $64$, and the weight decay is $1e-5$. For CIFAR10 and CIFAR100, we conduct experiments with ResNet-10~\cite{resnet8ju}, and for TinyImageNet and FC100, we use MobileNetV2~\cite{sandler2018mobilenetv2} and ProtoNet~\cite{protonet2017} respectively. The feature vector dimension is 512. The $M$ of FedSC in Eq.~\ref{eq:eq9} is set as 2, the $\tau$ in Eq.~\ref{eq:eq14} is set as 0.05. All experiments are run by PyTorch on two NVIDIA RTX A5000 GPUs.\\
\textbf{Baselines:} We compare ours against several SOTA FL methods: standard (FedAvg~\cite{McMahan17}, FedProx~\cite{li2020fedprox}); contrast-based (MOON~\cite{moon2021}, FedRCL~\cite{seo2024relaxed}); prototype-based (FedProto~\cite{tan2022fedproto}, FedNH~\cite{fednh2023}); and aggregation-based (FedNTD~\cite{lee2022preservation}, FedCDA~\cite{wang2024fedcda}).

\begin{table}[t]
\caption{Comparison results of convergence rates on CIFAR10 and CIFAR100. The $R_{acc}$ columns denote the minimum number of rounds required to reach $acc$ of the test accuracy, and the ACC columns show the final test accuracy.}
\label{tab:tab3}
\centering
\resizebox{\linewidth}{!}{ 
\begin{tabular}{l||cccc|cccc}
    \Xhline{1.2px}
    \rowcolor{mygray} & \multicolumn{4}{c}{\textbf{CIFAR10} ($\mathtt{NID1}_{0.2}$)} & \multicolumn{4}{c}{\textbf{CIFAR100} ($\mathtt{NID1}_{0.2}$)} \\
    \cline{2-9}
    \rowcolor{mygray} \multirow{-1.8}{*}{\textbf{Methods}} & $R_{20\%}$ & $R_{40\%}$ & $R_{80\%}$ & ACC & $R_{20\%}$ & $R_{40\%}$ & $R_{60\%}$ & ACC \\
    \hline\hline
    FedAvg & 5 & 18 & 55 & 84.65 & 9 & 35 & 79 & 60.08 \\
    FedProx & 5 & 15 & 52 & 84.59 & 9 & 32 & 76 & 60.21 \\
    MOON & 3 & 10 & 47 & 86.10 & 6 & 28 & 67 & 61.48 \\
    FedNTD & 4 & 12 & 47 & 86.23 & 7 & 30 & 65 & 61.74 \\
    FedNH & 4 & 13 & 50 & 85.25 & 10 & 32 & 73 & 60.43 \\
    FedRCL & 3 & 8 & 45 & 86.89 & 6 & 26 & 62 & \underline{62.36} \\
    FedCDA & 3 & 10 & 43 & \underline{87.11} & 7 & 26 & 65 & 61.95 \\
    \hline\hline
    \rowcolor{myye} FedSC & 2 & 6 & 36 & \textbf{88.13} & 5 & 20 & 53 & \textbf{63.25} \\
    \Xhline{1.2px}
\end{tabular}}
\end{table}

\begin{table}[t]
\caption{Ablation study of the key modules RPCL and CPDR.}
\label{tab:tab4}
\centering
\resizebox{\linewidth}{!}{ 
\begin{tabular}{c c||c c c c|c}
    \Xhline{1.2px}
    \rowcolor{mygray} &  & \multicolumn{5}{c}{\textbf{CIFAR10}} \\
    \cline{3-7}
    \rowcolor{mygray} \multirow{-1.8}{*}{\textit{RPCL}} & \multirow{-1.8}{*}{\textit{CPDR}} 
        & $\mathtt{NID1}_{0.05}$ & $\mathtt{NID1}_{0.2}$ & $\mathtt{NID1}_{0.5}$ & $\mathtt{NID2}$ & AVG \\
    \hline\hline
              &             & 78.27 & 84.65 & 86.11 & 75.60 & 81.15 \\
    \ding{51} &             & 81.17 & 86.79 & 88.63 & 78.39 & 83.74 \\
              & \ding{51}   & 80.35 & 85.20 & 88.02 & 77.41 & 82.75 \\
    \rowcolor{myye} \ding{51} & \ding{51} & 82.41 & 88.13 & 89.29 & 79.61 & \textbf{84.87} \\
    \hline\hline
    \rowcolor{mygray} &  & \multicolumn{5}{c}{\textbf{CIFAR10-LT}} \\
    \cline{3-7}
    \rowcolor{mygray} \multirow{-1.8}{*}{\textit{RPCL}} & \multirow{-1.8}{*}{\textit{CPDR}}
        & $\rho=10$ & $\rho=50$ & $\rho=100$ & $\rho=200$ & AVG \\
    \hline\hline
              &  & 75.94 & 63.25 & 52.97 & 49.69 & 60.46\\
    \ding{51} &  & 76.82 & 64.70 & 54.21 & 50.33 & 61.51\\
              & \ding{51} & 76.19 & 64.05 & 54.39 & 50.67 & 61.33\\
    \rowcolor{myye} \ding{51} & \ding{51} & 77.31 & 65.30 & 56.34 & 51.12 & \textbf{62.52}\\
    \Xhline{1.2px}
\end{tabular}}
\end{table}

\subsection{Evaluation Results}
\noindent \textbf{Performance Comparison:} The results for FedSC and compared FL baselines with different heterogeneity settings $\mathtt{NID1}$, $\mathtt{NID2}$, and imbalanced levels $\rho$ are shown in Table.~\ref{tab:tab1} and Table.~\ref{tab:tab2}. 
We observe that FedSC universally outperforms other methods, which implies that FedSC effectively unleash the client's semantic knowledge via prototypes to improve the model training. 
The experiments show that: 1) FedSC consistently outperforms others under different \textit{non-iid} settings, indicating that FedSC is robust to different heterogeneity levels, even in the more difficult $\mathtt{NID2}$ setting; 2) FedSC achieves significantly better results on imbalance settings with long-tail distribution, which confirms the benefit of regularization via consistent prototypes; 3) FedSC potentially enhances diversity in local training via relational prototypes, and yields improvements over baselines under the few-shot scenario.

\noindent \textbf{Convergence Rates:} The comparison results of convergence rates on CIFAR10 and CIFAR100 are shown in Table.~\ref{tab:tab3}. FedSC required fewer rounds to converge compared with others and achieved a higher final accuracy, showing the superiority of FedSC in dealing with \textit{non-iid} data and its robust stability, which also indicates that FedSC helps to promote training generalizability and efficiency. Note that due to the constraints of the proximal term, FedProx converges slightly faster than FedAvg in our experiments.

\begin{figure}[t]
\centering
    \begin{minipage}[t]{0.495\linewidth}
        \centering
        \includegraphics[width=1.0\linewidth]{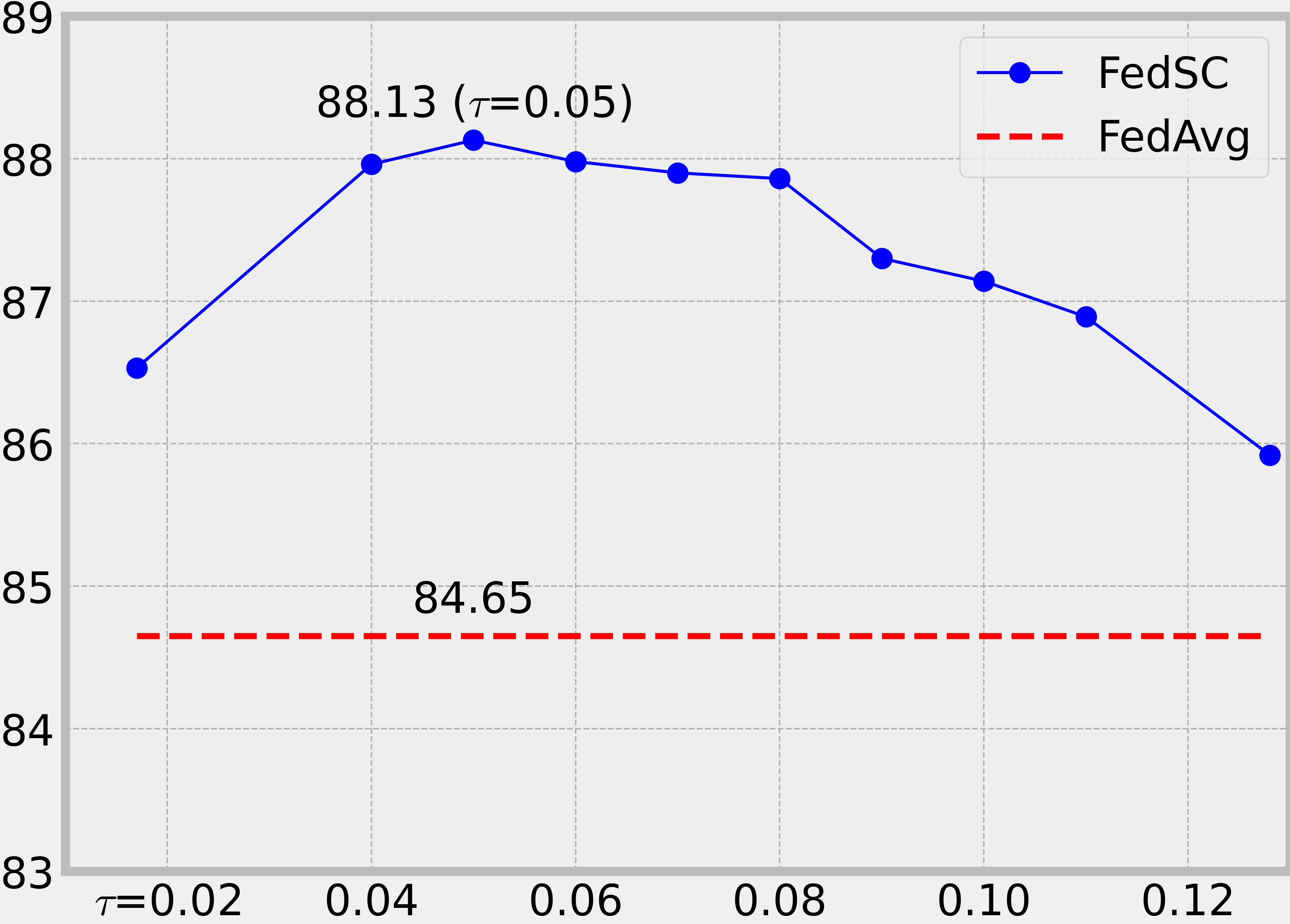}
    \end{minipage}
    \begin{minipage}[t]{0.495\linewidth}
        \centering
        \includegraphics[width=1.0\linewidth]{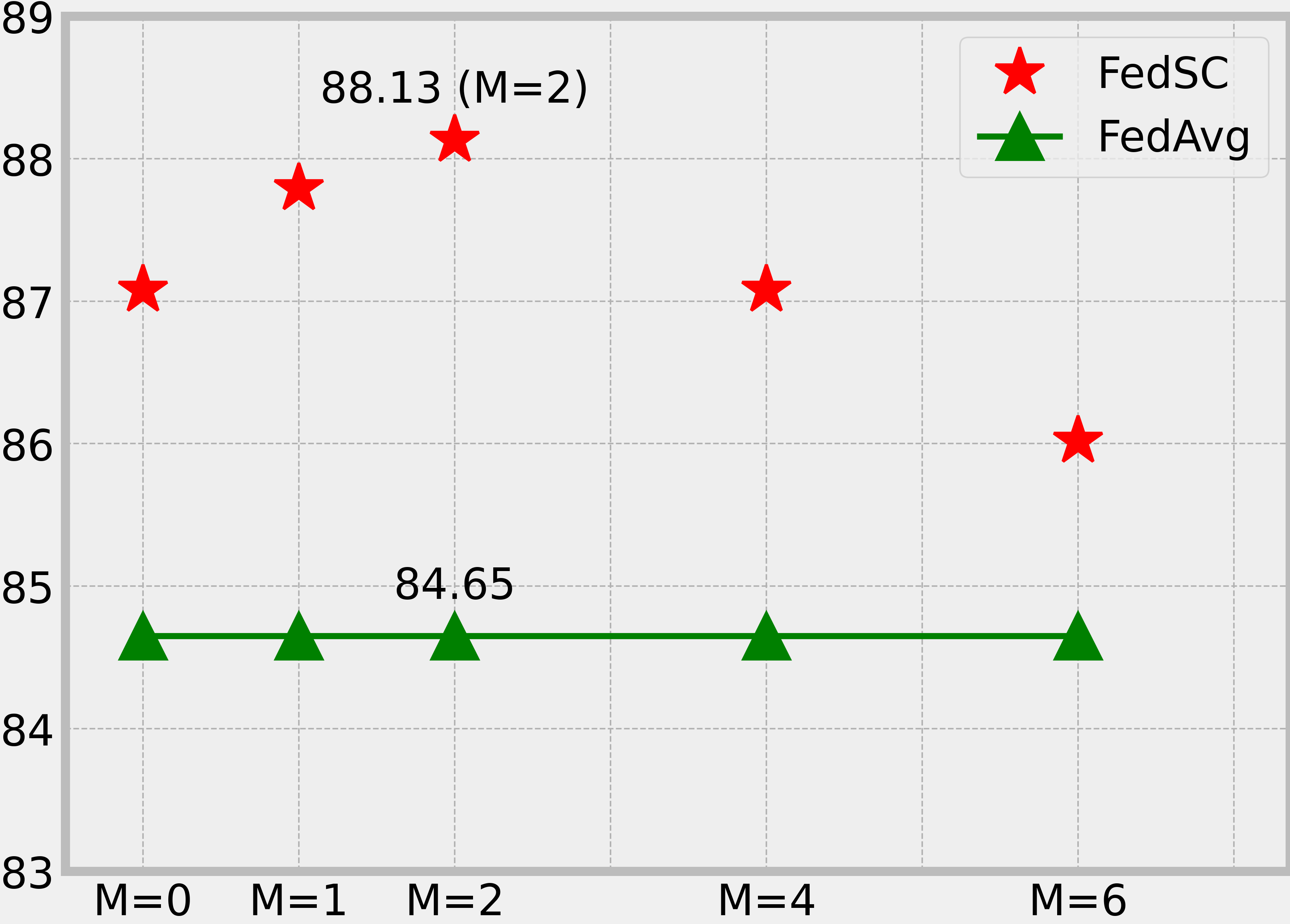}
    \end{minipage}
\caption{Effects of $\tau$ and $M$ on CIFAR10 dataset with $\mathtt{NID1}_{0.2}$.}
\label{fig:fig5}
\end{figure}

\begin{figure}[t]
\centering
    \begin{minipage}[t]{0.495\linewidth}
        \centering
        \includegraphics[width=1.0\linewidth]{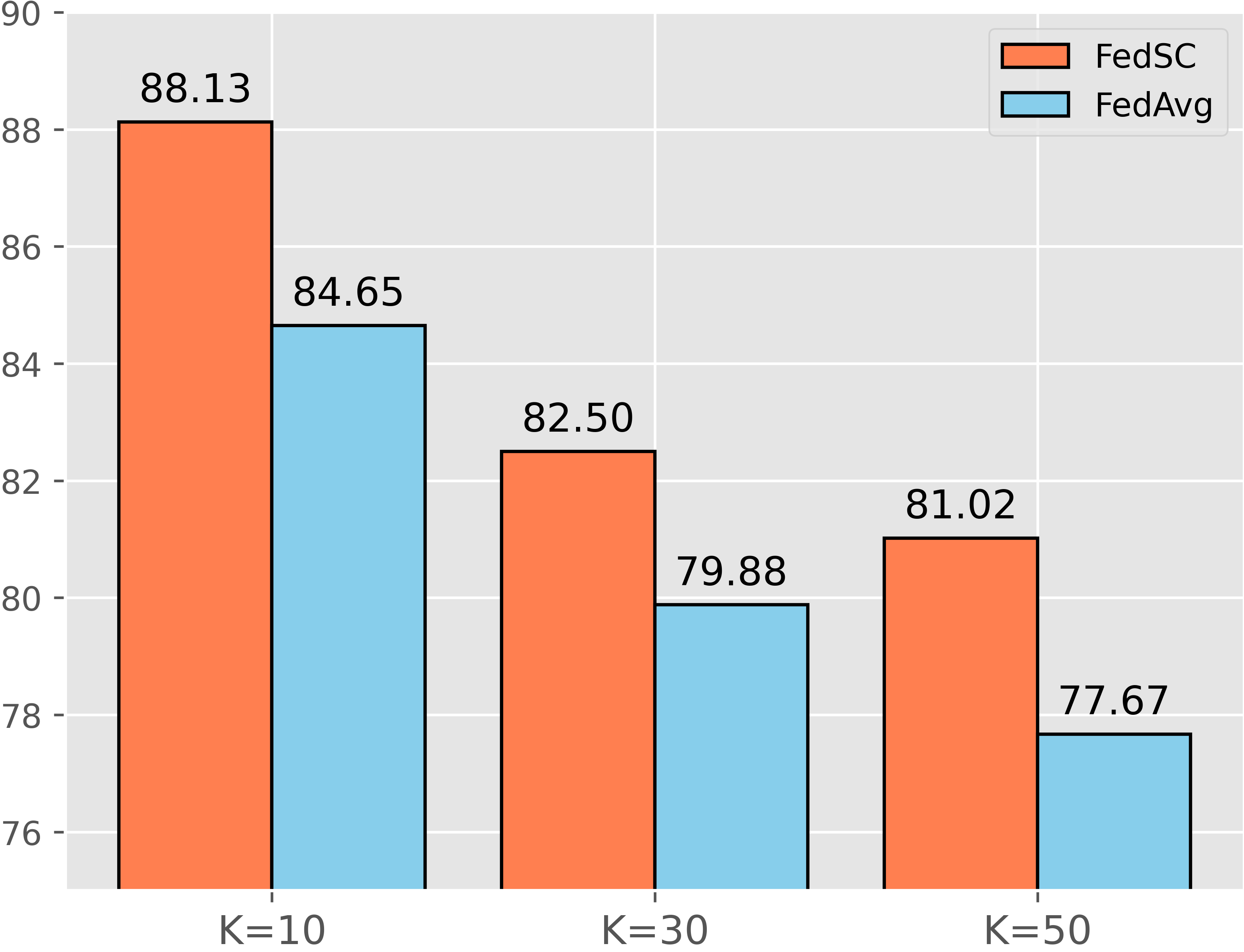}
    \end{minipage}
    \begin{minipage}[t]{0.495\linewidth}
        \centering
        \includegraphics[width=1.0\linewidth]{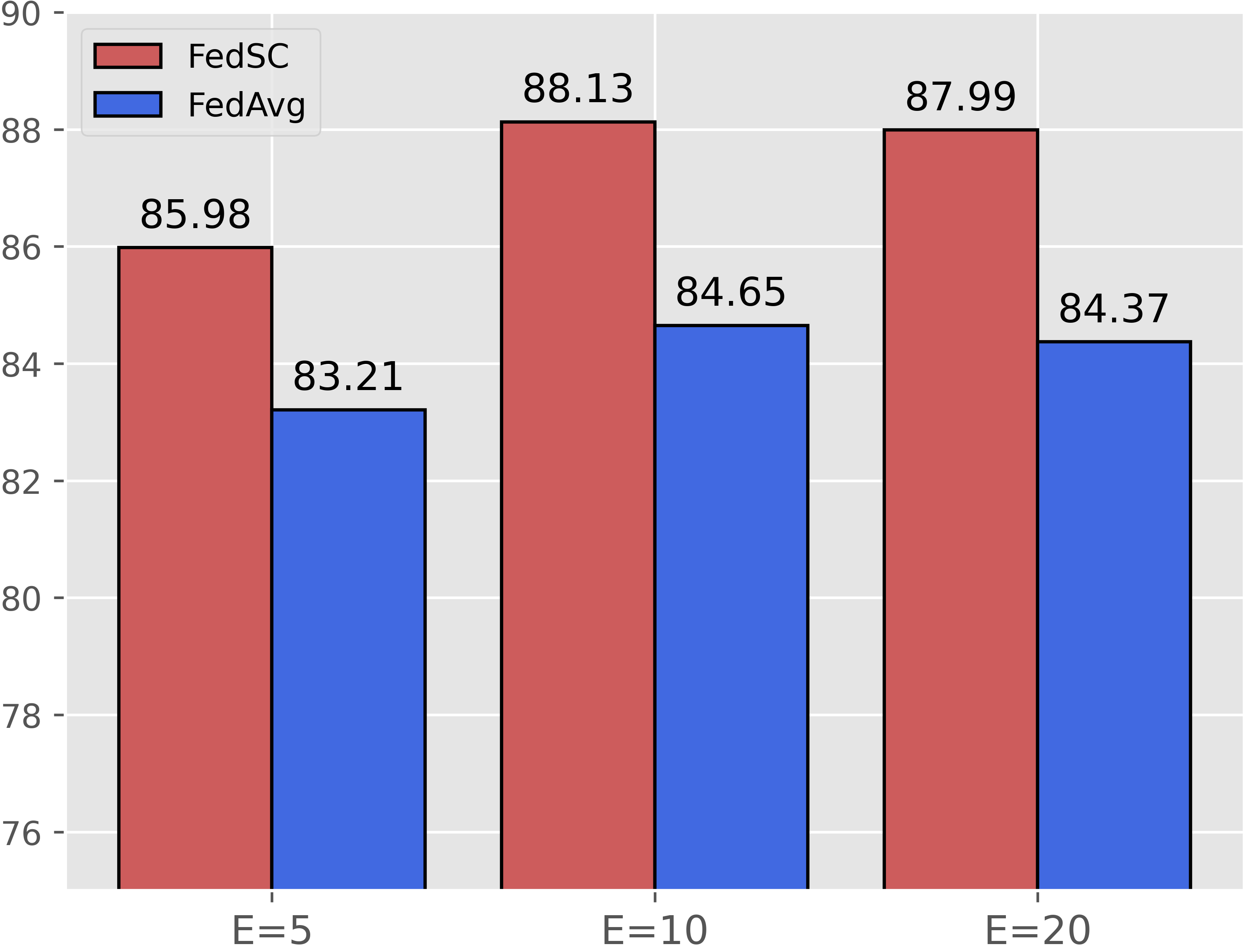}
    \end{minipage}
\caption{Ablation study on clients $K$ and local epochs $E$.}
\label{fig:fig6}
\end{figure}

\begin{table}[t]
\caption{Effects of selecting neighbor metrics in Eq.~\ref{eq:eq7}.}
\label{tab:tab5}
\centering
\begin{tabular}{l|ccccc}
\toprule   
    Metric & $L_{1}$ & $L_{2}$ & Cosine & KL Distance & Angle-wise \\
    \midrule
    ACC & 87.43 & 87.62 & 87.80 & 87.89 & \textbf{88.13} \\
\bottomrule
\end{tabular}
\end{table}

\begin{figure}[tbp]
    \centering
    \includegraphics[width=1.0\linewidth]{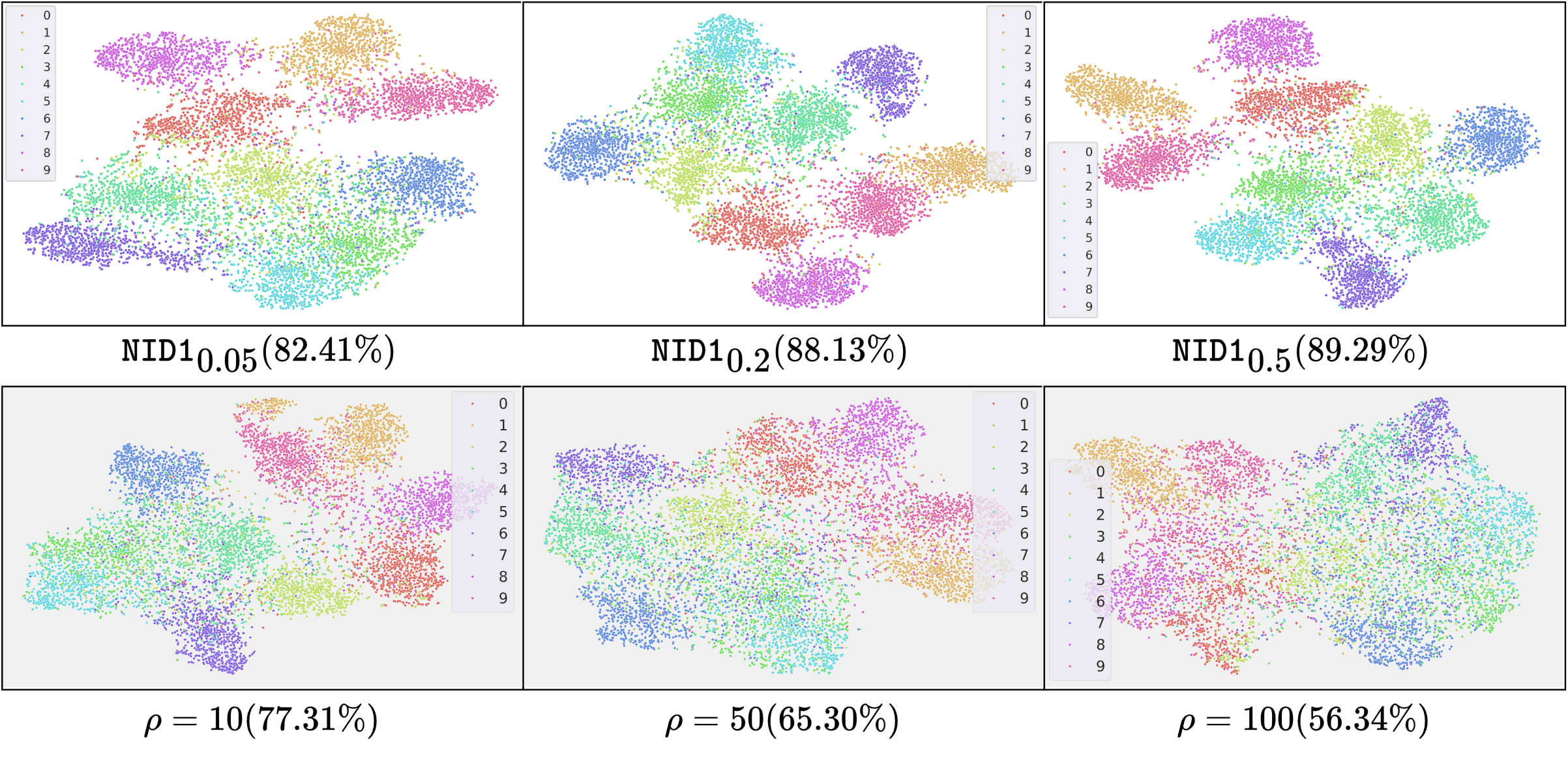}
    \caption{t-SNE visualization of FedSC on various scenarios.}
    \label{fig:fig7}
\end{figure}

\subsection{Validation Analysis}
To thoroughly analyze the efficacy of key modules, we perform an ablation study to investigate RPCL and CPDR modules. We give a quantitative result in Table.~\ref{tab:tab4}, the first row refers to FedAvg, which directly averages the model without extra operation: 
1) RPCL leads to significant performance improvements against the baseline, showing that RPCL is able to promote generalizable feature space; 
2) CPDR also yields solid gains, proving the importance of aligning instance embedding via regularization; 
3) combining RPCL and CPDR achieves better performance, supporting our motivation to unleash the inherent semantic knowledge via prototypes.

\noindent \textbf{Ablation Study:} In Fig.~\ref{fig:fig5}, we present the performance for different $\tau$ in Eq.~\ref{eq:eq14} and $M$ in Eq.~\ref{eq:eq9}, we observe that RPCL is not sensitive to the $\tau \in [0.04,0.08]$, showing the effectiveness of the normalization factor $\mathcal{U}$ in Eq.~\ref{eq:eq13}, capturing relative distances among sample pairs to match similarity. Besides, an excessively high $M$ leads to poor performance for FedSC, showing that information from other clients is not universally beneficial. FedSC reaches the optimal accuracy when $\tau=0.05$ and $M=2$. 
From Fig.~\ref{fig:fig6} Left, we see that FedSC brings significant performance gains even under more challenging settings with more clients. In the Fig.~\ref{fig:fig6} Right, FedSC consistently improves over the baselines across different epochs $E$. 
In Table.~\ref{tab:tab5}, we compare five discrepancy metrics under the $\mathtt{NID1}_{0.2}$ setting on CIFAR10, which shows that FedSC uses the angle-wise differences to select clients with similar global semantics on prototypes is more effective than these point-wise ways. 
We also plot the t-SNE visualization of FedSC in Fig.~\ref{fig:fig7}, depicting that FedSC can acquire well-generalizable ability during FL training under various scenarios.

\section{Conclusion}
In this paper, we explored the possibility of using intra-client semantically meaningful knowledge in handling data heterogeneity and proposed Federated Learning with Semantic-Aware Collaboration (FedSC). We construct relational prototypes and consistent prototypes at semantic-level, providing fruitful class-relevant knowledge and stable convergence signal. By leveraging the complementary advantages of two key modules RPCL and CPDR, FedSC promotes a more generalizable FL global model. The effectiveness of FedSC has been comprehensively analyzed from both the theoretical and experimental perspectives in various challenging FL scenarios.

\section*{Acknowledgments}
This work is supported by the scholarship from the China Scholarship Council (CSC) while the first author pursues his PhD degree in the University of Wollongong. This work was also partially supported by the Australian Research Council (ARC) Linkage Project under Grant LP210300009 and LP230100083.

\bibliographystyle{ACM-Reference-Format}
\bibliography{main}

\setcounter{table}{0}
\setcounter{figure}{0}
\renewcommand{\thefigure}{\Alph{section}.\arabic{figure}}
\renewcommand*{\theHfigure}{\thefigure}
\renewcommand{\thetable}{\Alph{section}.\arabic{table}}
\renewcommand*{\theHtable}{\thetable}
\setcounter{equation}{0}
\setcounter{theorem}{0}
\appendix

\section{Theoretical Analysis} \label{app:c}

\subsection{Preliminaries} \label{app:c1}
We express $\mathbf{e} \in \{1,2,\ldots,E\}$ as the local iteration step, $r$ as the global round, $E$ as the local epochs, $rE+\mathbf{e}$ refers to the $\mathbf{e}$-th local update in round $r+1$, $rE+1$ denotes the time between $r$-th global aggregation at the server and starting the model update on round $r+1$ at the local client.

\subsection{Assumptions} \label{app:c2}
Let $f_{k,r}(u_{k,r}) \in \mathbb{R}^{d}$ be the feature extractor to calculate client's prototypes, $h_{k,r}(v_{k,r}) \in \mathbb{R}^{|\mathcal{C}|}$ be the classifier to calculate logits, $w_{k,r}=\{u_{k,r}, v_{k,r}\}$ denotes the parameters of the model, $\mathcal{L}_{k,r}$ means the objective function, at the $k$-th client on the round $r$.

\begin{assumption}[Smoothness]
    \label{assumption1}
    For each objective function $\mathcal{L}$ is $L_{1}$-Lipschitz smooth, 
    which suggests the gradient of objective function $\mathcal{L}$ is $L_{1}$-Lipschitz continuous: 
    \begin{equation} \label{eq:app2}
    \small
    \begin{aligned}
        & \| \nabla\mathcal{L}_{k,r_{1}} - \nabla\mathcal{L}_{k,r_{2}}\|_{2} \leq 
        L_{1} \| w_{k,r_{1}} - w_{k,r_{2}}\|_{2}, \\
        & \forall \; r_{1},r_{2} > 0, \enspace k \in \{1,\ldots,K\},
    \end{aligned}
    \end{equation}
    since the smoothness depends on the derivative of the differentiable function, 
    it also indicates the following quadratic bound ($L_{1} > 0$):
    \begin{equation} \label{eq:app3}
    \small
    \begin{aligned}
        \mathcal{L}_{k,r_{1}} - \mathcal{L}_{k,r_{2}} \leq & 
        (\nabla\mathcal{L}_{k,r_{2}})^{\top} (w_{k,r_{1}} - w_{k,r_{2}}) \\
        & + \frac{L_{1}}{2} \| w_{k,r_{1}} - w_{k,r_{2}}\|_{2}^{2}.
    \end{aligned}
    \end{equation}
\end{assumption}

\begin{assumption}[Unbiased Gradient and Bounded Variance]
    \label{assumption2}
    For each client, the stochastic gradient is unbiased ($\zeta$ is the sample of the client dataset), 
    so the unbiased gradient is expressed as follows:
    \begin{equation} \label{eq:app4}
    \small
    \begin{aligned}
        & \mathbb{E}_{\zeta}[\nabla \mathcal{L}(w_{k,r}, \zeta)] = \nabla \mathcal{L}(w_{k,r}) = \nabla \mathcal{L}_{r}, \\
        & \forall \; k \in \{1,\ldots,K\}, \enspace r>0,
    \end{aligned}
    \end{equation}
    and has bounded variance: 
    \begin{equation} \label{eq:app5}
    \small
    \begin{aligned}
        & \mathbb{E}_{\zeta}[\| \nabla \mathcal{L}(w_{k,r}, \zeta) - \nabla \mathcal{L}(w_{k,r}) \|_{2}^{2}] \leq 
        \sigma^{2}, \\
        & \forall \; k \in \{1,\ldots,K\}, \enspace \sigma^{2} \geq 0.
    \end{aligned}
    \end{equation}
\end{assumption}

\begin{assumption}[Bounded Dissimilarity of Stochastic Gradients]
    \label{assumption3}
    For each objective function $\mathcal{L}$, 
    there exists constant $B$ such that the stochastic gradient is bounded: 
    \begin{equation} \label{eq:app6}
    \small
    \begin{aligned}
        & \mathbb{E}_{\zeta}[\| \nabla \mathcal{L}(w_{k,r}, \zeta) \|_{2}] \leq B, \\
        & \forall \; k \in \{1,\ldots,K\}, \enspace B>0, \enspace r>0.
    \end{aligned}
    \end{equation}
\end{assumption}

\begin{assumption}[Continuity]
    \label{assumption4}
    For each real-valued function $f(u)$ is $L_{2}$-Lipschitz continuous:
    \begin{equation} \label{eq:app7}
    \small
    \begin{aligned}
        & \| f_{k,r_{1}}(u_{k,r_{1}}) - f_{k,r_{2}}(u_{k,r_{2}}) \| \leq 
        L_{2} \| u_{k,r_{1}} - u_{k,r_{2}}\|_{2}, \\
        & \forall \; r_{1},r_{2} > 0, \enspace k \in \{1,\ldots,K\}.
    \end{aligned}
    \end{equation}
\end{assumption}

\subsection{Completing Proofs} \label{app:c3}
\subsubsection{Completing the Proof of Theorem~\ref{theorem1}}
\begin{proof} \label{proof1}
    We omit the client's notation $k$, and define the gradient descent as $w_{r+1} = w_{r} - \eta \nabla \mathcal{L}_{r}(w_{r}) = w_{r} - \eta \theta_{r}$, then we can get:
    \begin{equation}
    \small
        \begin{aligned}
            \mathcal{L}_{rE+2} 
            & \overset{(\romannumeral1)}{\leq} \mathcal{L}_{rE+1} + (\nabla\mathcal{L}_{rE+1})^{\top} 
                (w_{rE+2} - w_{rE+1}) \\
                & \quad \quad + \frac{L_{1}}{2} \| w_{rE+2} - w_{rE+1} \|_{2}^{2} \\
            & = \mathcal{L}_{rE+1} - \eta (\nabla\mathcal{L}_{rE+1})^{\top} \theta_{rE+1} + 
                \frac{L_{1}}{2} \| \eta \theta_{rE+1} \|_{2}^{2},
        \end{aligned}
    \end{equation}
    where $(\romannumeral1)$ from the $L_{1}$-Lipschitz quadratic bound of Assumption~\ref{assumption1}. 
    Then, we perform the expectation on both sides:
    \begin{equation}
    \small
        \begin{aligned}
            \mathbb{E}[\mathcal{L}_{rE+2}] 
            & \leq \mathcal{L}_{rE+1} - \eta \mathbb{E}[(\nabla\mathcal{L}_{rE+1})^{\top} \theta_{rE+1}] + 
                \frac{L_{1}}{2} \eta^{2} \mathbb{E}[\| \theta_{rE+1} \|_{2}^{2}] \\
            & \overset{(\romannumeral1)}{=} \mathcal{L}_{rE+1} - \eta \mathbb{E}[\| \nabla\mathcal{L}_{rE+1} \|_{2}^{2}] + 
                \frac{L_{1}}{2} \eta^{2} \mathbb{E}[\| \theta_{rE+1} \|_{2}^{2}] \\ 
            & \overset{(\romannumeral2)}{\leq} \mathcal{L}_{rE+1} - \eta \mathbb{E}[\| \nabla\mathcal{L}_{rE+1} \|_{2}^{2}] + 
                \frac{L_{1}}{2} \eta^{2} (\mathbb{E}^{2}[\| \nabla\mathcal{L}_{rE+1} \|_{2}] \\
                & \quad \quad + \textnormal{Variance} [\| \theta_{rE+1} \|_{2}] ) \\
            & = \mathcal{L}_{rE+1} - \eta \| \nabla\mathcal{L}_{rE+1} \|_{2}^{2} + \frac{L_{1}}{2} \eta^{2} 
                    (\| \nabla\mathcal{L}_{rE+1} \|_{2}^{2} \\
                & \quad \quad + \textnormal{Variance} [\| \theta_{rE+1} \|_{2}]) \\
            & = \mathcal{L}_{rE+1} + (\frac{L_{1}}{2} \eta^{2} - \eta) \| \nabla\mathcal{L}_{rE+1} \|_{2}^{2} \\
                & \quad \quad + \frac{L_{1}}{2} \eta^{2} \textnormal{Variance} [\| \theta_{rE+1} \|_{2}] \\
            & \overset{(\romannumeral3)}{\leq} \mathcal{L}_{rE+1} + (\frac{L_{1}}{2} \eta^{2} - \eta) 
                    \|\nabla\mathcal{L}_{rE+1}\|_{2}^{2} + \frac{L_{1}}{2} \eta^{2} \sigma^{2},
        \end{aligned}
    \end{equation}
    where $(\romannumeral1)$ from the stochastic gradient is unbiased in Eq.~\ref{eq:app4} of the Assumption~\ref{assumption2}; 
    $(\romannumeral2)$ from the formula $\textnormal{Variance}(x)=\mathbb{E}[x^{2}]-\mathbb{E}^{2}[x]$; 
    $(\romannumeral3)$ from the Eq.~\ref{eq:app5} of the Assumption~\ref{assumption2}. 
    Then, we telescope $E$ steps on both sides of the above equation:
    \begin{equation} \label{eq:app14}
    \small
        \begin{aligned}
            \mathbb{E}[\mathcal{L}_{(r+1)E}] 
            & \leq \mathcal{L}_{rE+1} + (\frac{L_{1}}{2} \eta^{2} - \eta) 
            \sum_{\mathbf{e}=1}^{E-1} \|\nabla\mathcal{L}_{rE+\mathbf{e}}\|_{2}^{2} + \frac{L_{1}}{2}E\eta^{2}\sigma^{2},
        \end{aligned}
    \end{equation}
    and then we consider the relationship between $\mathcal{L}_{rE+1}$ and $\mathcal{L}_{rE}$, 
    that is, given the any sample $\zeta$, we consider $\mathcal{L}_{(r+1)E+1}$ and $\mathcal{L}_{(r+1)E}$:
    \begin{equation*}
    \small
        \begin{aligned}
        \mathcal{L}_{(r+1)E+1}
            & = \mathcal{L}_{(r+1)E} + (\mathcal{L}_{CE,(r+1)E+1}+\mathcal{L}_{RPCL,(r+1)E+1} + \\
                & \quad \mathcal{L}_{CPDR,(r+1)E+1}) - \mathcal{L}_{(r+1)E} \\
            & = \mathcal{L}_{(r+1)E} + (\mathcal{L}_{RPCL,(r+1)E+1}+\mathcal{L}_{CPDR,(r+1)E+1}) - \\
                & \quad (\mathcal{L}_{RPCL,(r+1)E}+\mathcal{L}_{CPDR,(r+1)E}) \\ 
            & \overset{(\romannumeral1)}{\leq} \mathcal{L}_{(r+1)E} + 
                \log(\sum_{j=1}^{|\mathcal{C}|}(\text{exp}(s(\zeta,\mathcal{R}_{r+2}^{j})) - \text{exp}(s(\zeta,\mathcal{R}_{r+1}^{j})))) \\
                & \quad \quad + (\sum_{j=1}^{|\mathcal{C}|}(\|\mathcal{O}_{r+2}^{j}-\zeta \|_{2}-\|\mathcal{O}_{r+1}^{j}-\zeta\|_{2}))
        \end{aligned}
    \end{equation*}
    \begin{equation} \label{eq:app15}
    \small
    \begin{aligned}
            & \overset{(\romannumeral2)}{\leq} \mathcal{L}_{(r+1)E} + \sum_{j=1}^{|\mathcal{C}|}(
                    \| \mathcal{R}_{r+2}^{j} \|_{2} - \| \mathcal{R}_{r+1}^{j} \|_{2}) + \\
                & \quad \quad \sum_{j=1}^{|\mathcal{C}|}(\| \mathcal{O}_{r+2}^{j} \|_{2} - \| \mathcal{O}_{r+1}^{j} \|_{2}) \\
            & \overset{(\romannumeral3)}{\leq} \mathcal{L}_{(r+1)E} + \sum_{j=1}^{|\mathcal{C}|}((
                    \| \mathcal{R}_{r+2}^{j} - \mathcal{R}_{r+1}^{j} \|_{2})+(\| \mathcal{O}_{r+2}^{j} - \mathcal{O}_{r+1}^{j} \|_{2})) \\
            & \overset{(\romannumeral4)}{\leq} \mathcal{L}_{(r+1)E} + |\mathcal{C}|\sum_{k=1}^{K}\frac{n_{k}}{N}\frac{1}{M+1}(
                    \| f_{k,(r+1)E}(\zeta)-f_{k,rE}(\zeta) \|_{2}) \\
                & \quad \quad + |\mathcal{C}|\sum_{k=1}^{K}\frac{n_{k}}{N} (\| f_{k,(r+1)E}(\zeta)-f_{k,rE}(\zeta) \|_{2}) \\
            & = \mathcal{L}_{(r+1)E} + |\mathcal{C}|\sum_{k=1}^{K}\frac{n_{k}}{N}\frac{M+2}{M+1}(
                    \| f_{k,(r+1)E}(\zeta)-f_{k,rE}(\zeta) \|_{2}) \\
            & \overset{(\romannumeral5)}{\leq} \mathcal{L}_{(r+1)E}+
                |\mathcal{C}| L_{2} \sum_{k=1}^{K}\frac{n_{k}}{N}\frac{M+2}{M+1}(
                    \| u_{k,(r+1)E}-u_{k,rE} \|_{2}) \\
            & \overset{(\romannumeral6)}{\leq} \mathcal{L}_{(r+1)E}+
                |\mathcal{C}| L_{2} \sum_{k=1}^{K}\frac{n_{k}}{N}\frac{M+2}{M+1}(
                    \| w_{k,(r+1)E}-w_{k,rE} \|_{2}) \\
            & = \mathcal{L}_{(r+1)E}+|\mathcal{C}| L_{2} \sum_{k=1}^{K}\frac{n_{k}}{N}\frac{M+2}{M+1} 
                \| \eta \sum_{\mathbf{e}=1}^{E-1} \theta_{k,rE+\mathbf{e}}\|_{2} \\
            & = \mathcal{L}_{(r+1)E}+ \frac{|\mathcal{C}| L_{2} \eta (M+2)}{M+1} \sum_{k=1}^{K}\frac{n_{k}}{N} 
                \| \sum_{\mathbf{e}=1}^{E-1} \theta_{k,rE+\mathbf{e}} \|_{2} \\
            & \overset{(\romannumeral7)}{\leq} \mathcal{L}_{(r+1)E}+ \frac{|\mathcal{C}| L_{2} \eta (M+2)}
                    {M+1} \sum_{k=1}^{K}\frac{n_{k}}{N} 
                    \sum_{\mathbf{e}=1}^{E-1} \| \theta_{k,rE+\mathbf{e}} \|_{2},
    \end{aligned}
    \end{equation}
    where $(\romannumeral1)$ from the objective function in Eq.~\ref{eq:eq14} and Eq.~\ref{eq:eq15}; 
    $(\romannumeral2)$ from Eq.~\ref{eq:eq13}: $s(x,r^{j})=(\frac{1}{\mathcal{U}} \cdot \frac{x}{\|x\|_{2}} \cdot \frac{1}{\|r^{j}\|_{2}})r^{j} \leq r^{j}$, we take expectation to get $\mathbb{E}[s(x,r^{j})] \leq \|r^{j}\|_{2}$; 
    $(\romannumeral3)$ from the formula as $\|x\|_{2} - \|y\|_{2} \leq \|x-y\|_{2}$; 
    $(\romannumeral4)$ details from the Sec.~\ref{sec:rep} and Sec.~\ref{sec:cop}, and $\zeta$ is the sample of the client dataset; 
    $(\romannumeral5)$ from the Eq.~\ref{eq:app7} of the Assumption~\ref{assumption4}; 
    $(\romannumeral6)$ from the the fact that $u$ is a subset of $w=\{u, v\}$; 
    $(\romannumeral7)$ from the formula as $\| \sum{x} \|_{2} \leq \sum{\| x \|_{2}}$. 
    Then, we perform the expectation on both sides of the above equation Eq.~\ref{eq:app15}:
    \begin{equation}
    \small
        \begin{aligned}
            \mathbb{E}[\mathcal{L}_{(r+1)E+1}] 
            & \leq \mathcal{L}_{(r+1)E} + \frac{|\mathcal{C}| L_{2} \eta (M+2)}
                    {M+1} \sum_{k=1}^{K}\frac{n_{k}}{N} 
                    \sum_{\mathbf{e}=1}^{E-1} \mathbb{E}[\| \theta_{k,rE+\mathbf{e}} \|_{2}] \\
            & \overset{(\romannumeral1)}{\leq} \mathcal{L}_{(r+1)E} + \frac{|\mathcal{C}| L_{2} \eta EB (M+2)}{M+1},
        \end{aligned}
    \end{equation}
    where $(\romannumeral1)$ from the Eq.~\ref{eq:app6} of the Assumption~\ref{assumption3}. 
    Through the above equation, we can easily get:
    \begin{equation} \label{eq:app17}
    \small
        \mathbb{E}[\mathcal{L}_{rE+1}] \leq 
        \mathcal{L}_{rE} + \frac{|\mathcal{C}| L_{2} \eta EB (M+2)}{M+1},
    \end{equation}
    and then, based on Eq.~\ref{eq:app14} and Eq.~\ref{eq:app17}, take the expectation on both sides, we can get:
    \begin{equation} \label{eq:app18}
    \small
        \begin{aligned}
            \mathbb{E}[\mathcal{L}_{(r+1)E}]
            & \leq 
            \mathcal{L}_{rE} + \frac{|\mathcal{C}| L_{2} \eta EB (M+2)}{M+1} + \frac{L_{1}}{2}E\eta^{2}\sigma^{2} \\
                & \quad + (\frac{L_{1}}{2} \eta^{2} - \eta) 
                \sum_{\mathbf{e}=1}^{E-1} \|\nabla\mathcal{L}_{rE+\mathbf{e}}\|_{2}^{2} \\
            & \leq \mathcal{L}_{rE} - 
                (\eta - \frac{L_{1}}{2} \eta^{2})EB^{2} + 
                \frac{L_{1}E\eta^{2}}{2}\sigma^{2} \\
                & \quad + \frac{L_{2} E \eta |\mathcal{C}| B (M+2)}{M+1}.
        \end{aligned}
    \end{equation}
    
    Hence, we have completed the proof of Theorem~\ref{theorem1}.
\end{proof}

\subsubsection{Completing the Proof of Theorem~\ref{theorem2}}
\begin{proof}
    Based on the Theorem~\ref{theorem1}, we can easily get:
    \begin{equation}
    \small
        \mathbb{E}[\mathcal{L}_{(r+1)E}] - \mathcal{L}_{rE} \leq 
        \underbrace{\frac{(L_{1}\eta^{2} - 2\eta)EB^{2} + L_{1}E\eta^{2}\sigma^{2}}{2} + 
                \frac{L_{2} E \eta |\mathcal{C}| B (M+2)}{M+1}}_{T1}.
    \end{equation}
    Then, to ensure the right side of the above equation $T1 \leq 0$, we can easily get the following condition for $\eta$:
    \begin{equation} \label{eq:app20}
    \small
        \begin{aligned}
        \eta_{\mathbf{e}} < \frac{2(M+1)B^{2}-2(M+2)L_{2}|\mathcal{C}|B}{L_{1}(M+1)(\sigma^{2}+B^{2})}, \enspace \mathbf{e} \in \{1,2,\ldots,E-1\}.
        \end{aligned}
    \end{equation}
    So, the convergence of the objective function $\mathcal{L}$ in FedSC holds 
    as round $r$ increases and the above limitation of $\eta$ in Eq.~\ref{eq:app20}.
    
    Hence, we have completed the proof of Theorem~\ref{theorem2}.    
\end{proof}

\subsubsection{Completing the Proof of Theorem~\ref{theorem3}}
\begin{proof}
    Based on the Eq.~\ref{eq:app18}, we can get:
    \begin{equation}
    \small
        \begin{aligned}
            \mathbb{E}[\mathcal{L}_{(r+1)E}] 
            & \leq 
            \mathcal{L}_{rE} + \frac{|\mathcal{C}| L_{2} \eta EB(M+2)}{M+1} + \frac{L_{1}}{2}E\eta^{2}\sigma^{2} \\
            & \quad + (\frac{L_{1}}{2} \eta^{2} - \eta) 
            \sum_{\mathbf{e}=1}^{E-1} \|\nabla\mathcal{L}_{rE+\mathbf{e}}\|_{2}^{2}, 
        \end{aligned}
    \end{equation}
    and the following equation:
    \begin{equation}
    \small
        \begin{aligned}
            \sum_{\mathbf{e}=1}^{E-1} \|\nabla\mathcal{L}_{rE+\mathbf{e}}\|_{2}^{2} 
            & \leq \frac{2(\mathcal{L}_{rE}-\mathbb{E}[\mathcal{L}_{(r+1)E}]) + 
                        \frac{2|\mathcal{C}| L_{2} \eta EB(M+2)}{M+1} +
                        L_{1}E\eta^{2}\sigma^{2}}{2\eta - L_{1}\eta^{2}},
        \end{aligned}
    \end{equation}
    let the round $r$ from $0$ to $R-1$, step $\mathbf{e}$ from $1$ to $E$, we can easily get:
    \begin{equation}
    \small
        \begin{aligned}
            \frac{1}{RE} \sum_{r=0}^{R-1} \sum_{\mathbf{e}=1}^{E-1} 
            \mathbb{E}[\|\nabla\mathcal{L}_{rE+\mathbf{e}}\|_{2}^{2}] 
            & \leq \frac{1}{RE} 
                    \frac{2\sum_{r=0}^{R-1}(\mathcal{L}_{rE}-\mathbb{E}[\mathcal{L}_{(r+1)E}]) + T2}
                    {2\eta - L_{1}\eta^{2}}, \\
            \textnormal{where} \enspace T2 = & \frac{2|\mathcal{C}| L_{2} \eta REB(M+2)}{M+1} +
                                L_{1}RE\eta^{2}\sigma^{2},
        \end{aligned}
    \end{equation}
    and then, given any $\xi>0$, $\mathcal{L}$ and $R$ receive the following limitation:
    \begin{equation} \label{eq:app24}
    \small
        \frac{\frac{2}{RE}\sum_{r=0}^{R-1}(\mathcal{L}_{rE}-\mathbb{E}[\mathcal{L}_{(r+1)E}]) + 
                    \frac{2|\mathcal{C}| L_{2} \eta B (M+2)}{M+1} +
                    L_{1}\eta^{2}\sigma^{2}
                    }
        {2\eta - L_{1}\eta^{2}}
        < {\xi},
    \end{equation}
    according to the above equation in Eq.~\ref{eq:app24}, we can easily get the following condition for $R$:
    \begin{equation} \label{eq:app25}
    \small
        \begin{aligned}
        R > \frac{2(M+1)(\mathcal{L}_{0}-\mathcal{L}^{*})}{\underbrace{\xi E\eta(M+1)(2-L_{1}\eta)- (\varOmega_{1} + \varOmega_{2})}_{T3}},
        \end{aligned}
    \end{equation}
    where $\varOmega_{1}=(M+1)L_{1}E\eta^{2}\sigma^{2}$, $\varOmega_{2}=2(M+2)L_{2}E\eta|\mathcal{C}|B$, $\mathcal{L}^{*}$ denotes the optimal solution of $\mathcal{L}$, 
    and $\sum_{r=0}^{R-1}(\mathcal{L}_{rE}-\mathbb{E}[\mathcal{L}_{(r+1)E}])=
    \mathcal{L}_{0} - \mathcal{L}_{1} + \mathcal{L}_{1} - \mathcal{L}_{2} 
    + \cdots + \mathcal{L}_{R-1} - \mathcal{L}_{R} = \mathcal{L}_{0} - \mathcal{L}_{R} 
    \leq \mathcal{L}_{0} - \mathcal{L}^{*}$. 
    Then, the denominator of the above equation in Eq.~\ref{eq:app25} should satisfy $T3 > 0$, 
    so we can easily obtain the following condition for $\eta$:
    \begin{equation} \label{eq:app28}
    \small
    \eta < \frac{2 \xi (M+1) - 2(M+2)L_{2}|\mathcal{C}|B}
            {L_{1}(M+1)(\xi + \sigma^{2})}.
    \end{equation}
    Theorem~\ref{theorem3} provides the convergence rate of FedSC, which can confine the gradients of the objective function $\mathcal{L}$ to any bound, denoted as $\xi$, after carefully selecting the number of rounds $R$ in Eq.~\ref{eq:app25} and hyperparameters $\eta$ in Eq.~\ref{eq:app28}. The smaller $\xi$ is, the larger $R$ is, which means that the tighter the bound is, the more communication rounds $R$ is required.
\end{proof}

\end{document}